\documentclass[10pt,twocolumn,letterpaper]{article}

\usepackage[pagenumbers]{cvpr}

\usepackage{algorithm}
\usepackage{algpseudocode}
\usepackage[most]{tcolorbox}

\definecolor{cGraphA}{HTML}{5B7FA6}
\definecolor{cPriorA}{HTML}{6FA82F}
\definecolor{cFlowA}{HTML}{D9822B}
\definecolor{cAppA}{HTML}{8A7AB8}

\tcbset{
  algblock/.style={
    enhanced, colback=#1!10, colframe=#1,
    boxrule=0pt, leftrule=2.2pt, arc=0pt,
    left=3pt, right=3pt, top=1pt, bottom=1pt,
    boxsep=1pt, before skip=3pt, after skip=3pt,
    fonttitle=\footnotesize\bfseries,
    coltitle=#1!70!black,
    attach title to upper,
    after title={\par\vspace{1pt}}
  }
}

\algrenewcommand\alglinenumber[1]{%
  \footnotesize\color{gray}#1}

\usepackage{xcolor}
\usepackage{multirow}
\usepackage{booktabs}
\usepackage[normalem]{ulem}

\definecolor{cgraph}{HTML}{0072B2}
\definecolor{cprior}{HTML}{D55E00}
\definecolor{cflow}{HTML}{009E73}

\usepackage[table]{xcolor}
\definecolor{oursbg}{RGB}{233,242,252}
\newcommand{\sd}[1]{\,{\scriptsize\textcolor{gray}{$\pm$#1}}}

\usepackage{booktabs}
\usepackage{colortbl}
\usepackage{pifont}
\definecolor{oursrow}{RGB}{229,239,252}
\usepackage{adjustbox}

\definecolor{cAff}{HTML}{C7C7C7}
\definecolor{cSVF}{HTML}{E69F00}
\definecolor{cEla}{HTML}{A88BC7}
\definecolor{cOurs}{HTML}{009E73}
\definecolor{cComb}{HTML}{56B4E9}

\definecolor{allbg}{HTML}{E0F9FA}

\newsavebox{\dicetab}

\usepackage{currfile} 
\usepackage{etoc}
\usepackage{caption} 

\definecolor{cvprblue}{rgb}{0.21,0.49,0.74}
\usepackage[
    pagebackref,
    breaklinks,
    colorlinks,
    allcolors=cvprblue
]{hyperref}

\def\confName{CVPR}
\def\confYear{2027}

\title{Fast Anatomy-Aware Diffeomorphic Augmentation for Brain MRI}

\author{%
Tonmoy Hossain$^{1}$ \quad
Alexandra G. Roberts$^{1}$ \quad
Mark Olchanyi$^{1,2}$ \quad
Jonathan Williams Ramirez$^{1}$ \\
Stefan Brandst\"{a}tter$^{1}$ \quad
Juan Eugenio Iglesias$^{1,2,3}$ \\[0.6em]
$^{1}$Harvard Medical School and Massachusetts General Hospital \\
$^{2}$Massachusetts Institute of Technology\\
$^{3}$University College London
}

\begin{document}

\addtocontents{toc}{\protect\setcounter{tocdepth}{-1}}


\twocolumn[{%
\renewcommand\twocolumn[1][]{#1}%
\maketitle

\begin{center}
    \centering
    \includegraphics[width=0.98\textwidth, trim=0 0 0 0cm, clip]{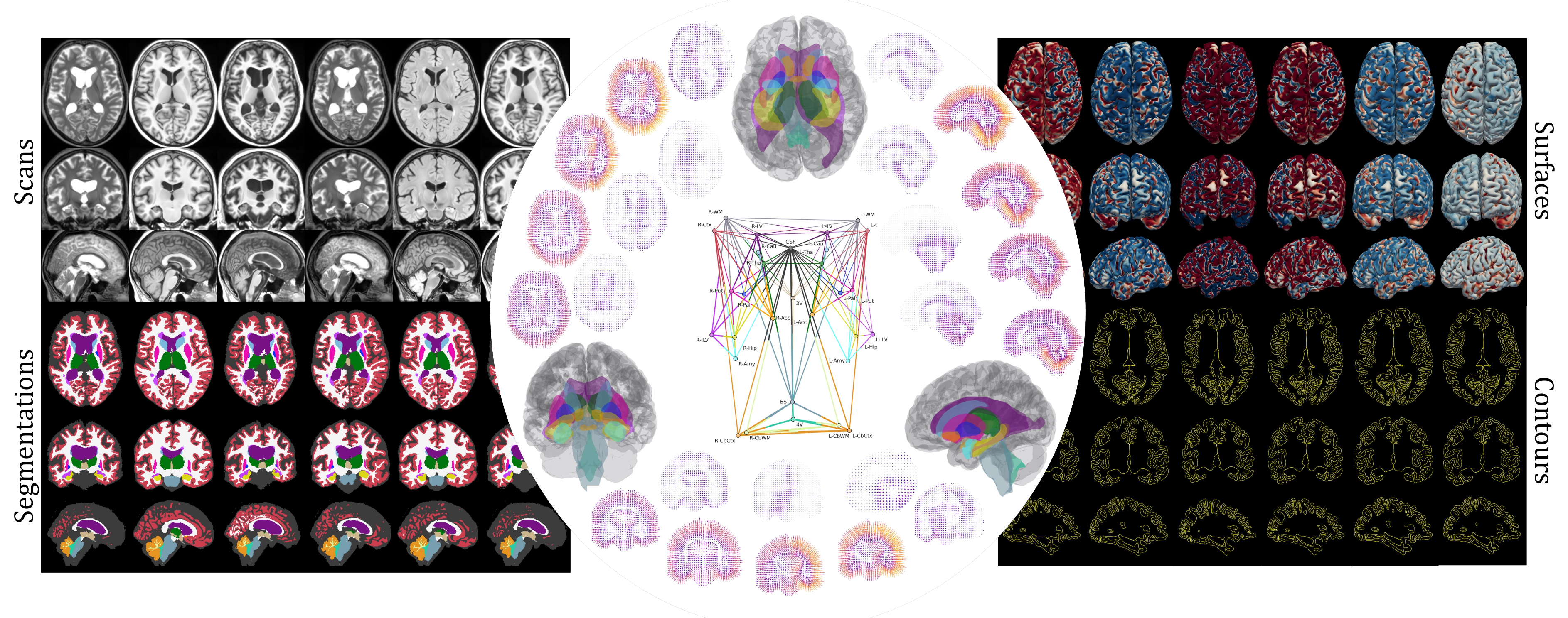}

    \captionsetup{hypcap=false}
    \captionof{figure}{\textbf{Anatomically coordinated augmentation from a single reference.} From a single brain MRI label map (center, three views), BrainAug constructs an anatomical graph encoding spatial proximity and bilateral correspondence between brain structures. A graph-based probabilistic model samples correlated regional volume changes to generate diverse, anatomically coordinated diffeomorphic deformations (surrounding circle). Applied to corresponding brain MRI scans, segmentations, cortical surfaces, and contours, these deformations produce spatially consistent anatomical variants (shown column-wise). Surface colors indicate local expansion (\textcolor{red}{red}) and contraction (\textcolor{blue}{blue}).}
    \captionsetup{hypcap=true}
    \label{fig:\currfilebase}%
\end{center}

\vspace{1em}
}]
\begin{abstract}
Data augmentation is essential for training deep learning models that are generalizable, particularly when training data are limited. In brain MRI, augmentation methods that synthesize anatomically diverse images without relying on large cohorts are particularly attractive, especially when training supervised models for scans with a specific resolution or MRI contrast. While conventional random nonlinear fields offer a simple means of achieving such diversity, they fail to capture the underlying spatial dependencies among morphological changes across brain regions. To address this limitation, we introduce \texttt{BrainAug}, an anatomy-aware augmentation framework that generates diverse, anatomically coordinated deformations without relying on population imaging data. \texttt{BrainAug} constructs an anatomical graph encoding spatial proximity and bilateral correspondence between brain regions, then incorporates these relationships into a probabilistic model designed to sample correlated regional volume changes. These sampled variations are translated into diffeomorphic transformations, enabling substantial structural changes while preserving anatomical topology. Combined with contrast-agnostic image synthesis, \texttt{BrainAug} generates training images with independently varying anatomy and appearance. Evaluations across eight brain MRI datasets demonstrate stronger segmentation generalization across diverse populations, while registration experiments show improved alignment under substantial morphological differences. 
\end{abstract}
\section{Introduction}
\label{sec:intro}

Data augmentation plays a crucial role in training deep learning models that generalize across diverse brain anatomies, including variations arising from development, aging, and disease progression~\cite{bethlehem2022brain}. Achieving such generalization requires exposing models to a broad range of anatomical variations and imaging conditions that may not be adequately represented in the available training data. Despite the growing availability of large-scale neuroimaging datasets, obtaining images that capture sufficient anatomical diversity while meeting specific contrast and resolution requirements remains challenging~\cite{wachinger2021detect,bento2022deep}. Augmentation thus offers a way to broaden this exposure beyond the constraints of available imaging data, facilitating the training of robust deep networks.

Spatial augmentation provides a practical means of introducing anatomical variability into training data. Conventional approaches use random affine and smooth deformable transformations to diversify brain anatomy, offering a simple but efficient way to capture anatomical variability without requiring additional imaging cohorts or model training. These anatomy-agnostic spatial augmentations are commonly combined with intensity augmentations, such as brightness, contrast, gamma correction, and resolution variations, to simulate changes in image appearance. More recently, the \emph{Synth}-family of frameworks has improved the generalizability of brain MRI analysis models through domain randomization, synthesizing images by drawing voxel intensities from label-conditioned Gaussian distributions, at the expense of accurately modeling regions with non-Gaussian intensity profiles~\cite{billot2023synthseg,iglesias2023synthsr,hoffmann2021synthmorph}.

While geometric augmentation effectively diversifies training anatomy, it does not explicitly model relationships among regional changes. Anatomical structure and inter-dependencies play a critical role in coordinated morphological changes, such as regional tissue atrophy accompanied by enlargement of surrounding pockets of cerebrospinal fluid (CSF)~\cite{thompson2004mapping,apostolova2012hippocampal,iglesias2016bayesian}. A central challenge is therefore  generating diverse anatomical variations that reflect these dependencies. Biophysical and atrophy simulation approaches~\cite{karaccali2006simulation,camara2006phenomenological,khanal2016biophysical,khanal2017simulating} generate such changes from pre-specified regional atrophy patterns or those estimated from longitudinal images. While these methods can generate anatomically accurate variations, their computational cost makes them impractical for on-the-fly augmentation during network training, even with GPU acceleration.

Learning-based approaches offer a practical alternative by sampling augmentations from deformation distributions learned from a given set of images, using generative models~\cite{pombo2023equitable,hossain2025mgaug}, adversarial models~\cite{olut2020adversarial}, or task-driven approaches~\cite{chaitanya2021semi}, including methods that leverage registration to unlabeled images~\cite{zhao2019data,zhang2024one,shen2020anatomical,wang2020lt}. These approaches can capture complex deformation patterns that are seldom sampled by conventional anatomy-independent augmentation. However, these approaches rely on labeled/unlabeled imaging cohorts, which may be difficult to obtain at the scale needed for training deep networks with specific imaging contrasts and resolutions.

Here we introduce \texttt{BrainAug}, a fast, anatomy-aware augmentation framework that generates diverse deformations by modeling interdependent morphological changes across brain regions. Combined with contrast-agnostic image synthesis, these transformations enable the generation of training images with diverse brain anatomy under varying imaging appearances. Without requiring large cohorts or model training, \texttt{BrainAug} supports efficient on-the-fly augmentation and can be seamlessly incorporated into existing training pipelines. Our contributions are threefold:

\begin{itemize}
    \item Introduce a graph-based probabilistic model of regional volume changes that captures spatial and bilateral anatomical dependencies, enabling coordinated morphological variations that require neither population imaging data nor neural network training.
    \item Develop an efficient deformation generation framework that translates sampled volume changes into diffeomorphic transformations with approximate intracranial volume conservation, supporting variations from subtle regional changes to large anatomical differences.
    \item Demonstrate improved segmentation and registration generalization across diverse brain MRI datasets and populations, without target-dataset training or fine-tuning.
\end{itemize}


\section{Anatomy-Aware Deformation Generation}
\label{sec:ourModel}

\noindent \textbf{Problem Overview.}
Given an anatomical label map $S:\Omega\rightarrow\{0,\ldots,K\}$, containing $K$ distinct neuroanatomical labels and a background class ($0$), our goal is to synthesize diverse anatomical variations while preserving structural relationships among brain regions. \texttt{BrainAug} achieves this by modeling coordinated regional volume changes, enabling augmentation across different imaging contrasts and resolutions from a single reference anatomy without learning deformation distributions from training data.

\begin{figure}[htbp]
    \centering
    \includegraphics[width=\linewidth]{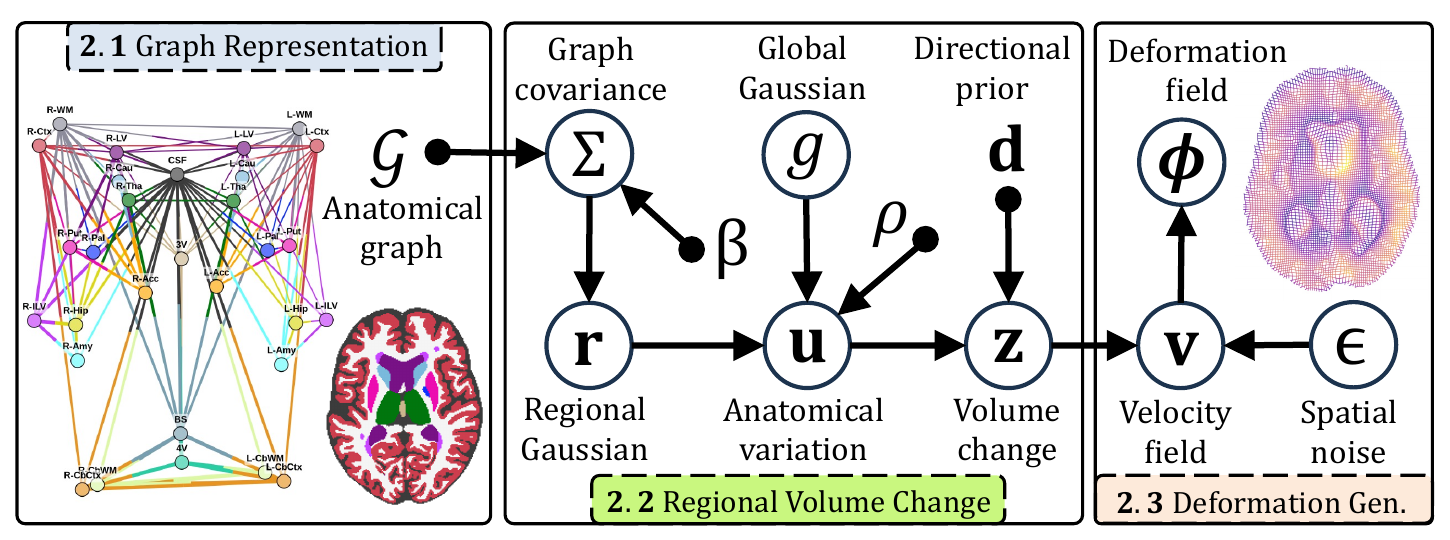}
    \caption{\textbf{Overview of \texttt{BrainAug}.} Anatomy-aware augmentation through graph-based modeling of correlated regional volume changes and their translation into diffeomorphic deformations.}
    \label{fig:graphRep}
\end{figure}

\texttt{BrainAug} (Fig.~\ref{fig:graphRep}) begins by constructing an anatomical graph $\mathcal{G}$ encoding spatial proximity and bilateral relationships among brain structures (Sec.~\ref{sec:graph}), a one-time operation for each input label map. Leveraging this representation, a probabilistic generative model samples correlated regional log-volume changes to capture coordinated expansion and contraction (Sec.~\ref{sec:prior}). To translate these volume changes into spatial deformations, we construct a divergence field that accounts for volume compensation between brain tissue and CSF. Solving a Poisson equation then yields a dense stationary velocity field (SVF), which is subsequently integrated to produce a diffeomorphic deformation $\phi:\Omega\rightarrow\Omega$ (Sec.~\ref{sec:flow}). The resulting deformation can be applied consistently to images and anatomical shapes, and combined with appearance and acquisition transformations to diversify training data (Sec.~\ref{sec:synth}). The entire augmentation process takes only $\mathbf{79.7\scriptstyle\,\pm2.1}$\,\textbf{ms} for a $182 \times 218 \times 182$-voxel volume on a single NVIDIA RTX PRO 6000 GPU.

\subsection{Region Dependency Graph Representation}
\label{sec:graph}
We construct a weighted anatomical graph $\mathcal{G}=(\mathcal{V},\mathcal{E},\mathcal{A})$ from the input label map $S$ to capture spatial and bilateral relationships among brain structures. Each node $n_i\in\mathcal{V}$ represents an anatomical region $\Omega_i=\{x\in\Omega\mid S(x)=i\}$, while $\mathcal{E}$ denotes the edges representing spatial and bilateral relationships between regions. The weighted adjacency matrix $\mathcal{A}\in\mathbb{R}^{K\times K}$ encodes these relationships and their corresponding weights through two components:
\begin{equation}
\mathcal{A}=\mathcal{A}^{\mathrm{spatial}}
+\lambda\mathcal{A}^{\mathrm{bilateral}}.
\end{equation}

The spatial component connects regions within a distance threshold $\tau$, with weights inversely proportional to the minimum distance $d_{ij}$ between their boundaries (in mm):

\begin{equation}
\mathcal{A}_{ij}^{\mathrm{spatial}} =
\begin{cases}
\displaystyle\frac{1}{1+d_{ij}}, & \text{if } d_{ij}\leq\tau,\\
0, & \text{otherwise}.
\end{cases}
\label{eq:graph_g}
\end{equation}

The bilateral component connects homologous structures across the left and right hemispheres, regardless of their spatial distance. We define $\mathcal{A}^{\mathrm{bilateral}}_{ij}=1$ for corresponding left--right regions and $0$ otherwise, with $\lambda=0.5\max_{i,j}\mathcal{A}^{\mathrm{spatial}}_{ij}$ scaling bilateral connections relative to spatial adjacency. Note that bilateral coupling is optional and can be disabled by setting $\lambda=0$. The resulting graph serves as the anatomical foundation for our probabilistic generative model, which captures inter-regional dependencies to synthesize coordinated and diverse brain volume changes. \textit{Importantly, this graph is constructed only once per input label map, with its adjacency matrix cached and reused during on-the-fly augmentation throughout training.}

\subsection{Probabilistic Model of Regional Volume Changes}
\label{sec:prior}

Building on the anatomical graph $\mathcal{G}$, we introduce a probabilistic generative model that captures inter-regional dependencies to sample coordinated volume changes across brain structures. To represent these changes, we define the regional log-volume ratio $z_i=\log(V_i'/V_i)$, where $V_i$ and $V_i'$ denote the original and target volumes of region $i$, respectively. This formulation models regional expansion ($z_i>0$) and contraction ($z_i<0$) symmetrically in log-volume space, assigning equal likelihood to reciprocal volume changes (e.g., halving and doubling).

To model inter-regional dependencies, we construct a graph-based covariance matrix $\Sigma_{\mathcal G}$ that governs correlations among regional volume variations. We derive it from the graph Laplacian $L=D-\mathcal{A}$, where $D_{ii}=\sum_j \mathcal{A}_{ij}$. Specifically, we construct a heat-kernel covariance $K_\beta=\exp(-\beta L)$ and normalize its diagonal to obtain:

\begin{equation}
\Sigma_{\mathcal G}
= D_K^{-1/2}K_\beta D_K^{-1/2},
\qquad
D_K=\operatorname{diag}(K_\beta),
\label{eq:volChange}
\end{equation}
where $\beta\geq 0$ controls correlations across the anatomical graph. This normalization maintains unit marginal variance, allowing $\beta$ to control inter-regional dependencies independently of the sampling magnitude.

Anatomical variations can involve both correlated changes among related structures and widespread changes across the brain. We therefore model these complementary patterns through graph-correlated regional and shared global components. To generate these variations, we sample a regional latent vector $\mathbf r\in\mathbb{R}^K$ from a multivariate Gaussian distribution and an independent global latent variable $g\in\mathbb{R}$ from a standard Gaussian distribution:
\begin{equation}
\mathbf r\sim\mathcal{N}(\mathbf 0,\Sigma_{\mathcal G}),
\qquad
g\sim\mathcal{N}(0,1).
\end{equation}
Here, $\mathbf r$ models graph-correlated regional variations, while $g$ represents a shared brain-wide component. We then combine these two sources of variation to define the latent anatomical variation vector $\mathbf u\in\mathbb{R}^K$:

\begin{equation}
\mathbf u=\sqrt{\rho}\,g\mathbf1+
\sqrt{1-\rho}\,\mathbf r,
\label{eq:corr}
\end{equation}
where $\mathbf 1\in\mathbb R^K$ is the all-ones vector and $\rho\in[0,1]$ controls the relative contributions of global and regional components to the variance while preserving unit marginal variance. Since $\mathbf r$ and $g$ are independent Gaussian variables, their linear combination $\mathbf u$ also follows a multivariate Gaussian distribution, $\mathbf u\sim\mathcal N(\mathbf0,\Sigma_u)$, with covariance:

\begin{equation}
\Sigma_u = \underbrace{\rho\mathbf{1}\mathbf{1}^\top}_{\text{Global dependence}} + \underbrace{(1-\rho)\Sigma_{\mathcal G}}_{\text{Anatomical dependence}}.
\end{equation}

This covariance decomposition explicitly captures brain-wide variations and positive dependencies among anatomically related regions, while preserving the unit marginal variance across structures. Opposing tissue and CSF volume changes are induced by the volume-compensation constraint (Sec.~\ref{sec:flow}). To optionally guide anatomical variations toward specified patterns of regional expansion and contraction, we introduce a directional vector $\mathbf d=(d_1,\ldots,d_K)$, where each $d_i$ specifies the direction and magnitude of regional change. We then transform the latent variables into bounded regional log-volume targets:

\begin{equation}
z_i=c\,\tanh\left(
\frac{\alpha d_i+\gamma u_i}{c}
\right),
\end{equation}
where $\alpha$ controls the strength of the directional pattern, $\gamma$ controls the magnitude of stochastic variations, and $c$ bounds the log-volume changes. By varying these parameters, the model produces diverse patterns of coordinated anatomical changes without learning a deformation distribution from, or requiring access to population MRI data.

\noindent \textbf{Flexible Anatomical Variation.} We support three bilateral sampling modes: (i) \emph{symmetric}: enforcing latent variations across hemispheres ($u_R=u_L$); (ii) \emph{asymmetric}: retaining sampled left--right variations without enforcing symmetry; and (iii) \emph{mixed}: randomly applying symmetric sampling to a subset of bilateral pairs. 

\subsection{Volume-constrained Deformation Generation}
\label{sec:flow}

Given the sampled regional log-volume changes $\mathbf z=(z_1,\ldots,z_K)$, we construct a dense deformation field that realizes these volumetric variations while maintaining spatial coherence across anatomical structures. We generate this deformation using a spatial divergence field, which governs local tissue expansion and contraction, similar to prescribed-atrophy simulation~\cite{khanal2016biophysical,khanal2017simulating,lorenzi2015regional}. Unlike these simulators, which solve a biophysical model with skull boundary conditions for a user-specified map, we use a lightweight curl-free Poisson formulation that can be solved efficiently, thus supporting on-the-fly generation when training neural networks.

\noindent \textbf{Volume-Constrained Divergence Field.} To introduce local anatomical variability, we generate a spatially correlated random field $\epsilon(x)$ by interpolating i.i.d.\ Gaussian noise with variance $\sigma_\epsilon^2$ sampled on a coarse grid. We then map the sampled regional changes to a voxel-wise divergence target, exploiting the approximate correspondence between velocity divergence and local log-volume change:

\begin{equation}
  f(x) = \hat{z}_{S(x)} + \epsilon(x),
  \label{eq:epsilon}
\end{equation}
where $\hat{z}_i$ denotes the calibrated regional target, obtained by scaling $z_i$ with a per-region factor. Since deformation integration (described below) only approximately realizes the prescribed regional volume changes, these factors are estimated once per label map from sampled deformations to improve the agreement between prescribed and realized volume changes. Exact correspondence is not required, as the objective is to generate diverse anatomical variations rather than precisely match target volumes.

To balance tissue volume changes, CSF voxels $\mathcal C$ are assigned a uniform divergence that offsets the net change of tissue voxels $\mathcal T$, $f_{\mathrm{csf}}=-\frac{1}{|\mathcal C|}\sum_{\mathbf y\in\mathcal T} f(\mathbf y)$, so that $f$ sums to zero over the brain and intracranial volume is approximately conserved.

\noindent \textbf{Poisson-based Velocity Construction.} We construct a stationary velocity field $\mathbf v:\Omega\rightarrow\mathbb R^3$ whose divergence approximates the prescribed divergence target
\begin{equation}
\nabla\cdot\mathbf v(\mathbf x)\approx f(\mathbf x).
\end{equation}

By expressing the velocity as the gradient of a scalar potential, $\mathbf v=\nabla\psi$, this reduces to a Poisson equation 

\begin{equation}
\Delta\psi(\mathbf x)=f(\mathbf x),
\qquad
\mathbf v(\mathbf x)=\nabla\psi(\mathbf x).
\end{equation}
We solve this equation in the Fourier domain on a zero-padded grid to efficiently obtain a dense, spatially smooth velocity field. The zero-frequency component is excluded to satisfy the compatibility condition of the solver; since $f$ vanishes outside the brain and sums to zero within it, this leaves the prescribed targets unchanged. This formulation translates the sampled regional volume changes into a continuous deformation-driving field without independently deforming anatomical structures.

\noindent \textbf{Deformation Integration.}
We obtain the pull-back deformation by inverse
integration of the SVF using $7$ scaling-and-squaring steps~\cite{arsigny2006log}:

\begin{equation}
\boldsymbol{\phi} = \exp(-\mathbf{v}), \qquad
\frac{\partial \boldsymbol{\phi}_t(\mathbf{x})}{\partial t}
= -\mathbf{v}\big(\boldsymbol{\phi}_t(\mathbf{x})\big),
\quad
\boldsymbol{\phi}_0(\mathbf{x}) = \mathbf{x}.
\end{equation}
where $\boldsymbol{\phi}=\boldsymbol{\phi}_1$ denotes
the pull-back deformation at $t=1$.

The resulting deformation is applied to the input image $X$ through the pull-back operation $X_\phi = X \circ \boldsymbol{\phi}$, and consistently to its label map to preserve spatial correspondence. These spatial deformations can be combined with intensity augmentations to generate anatomically diverse training images under varying imaging conditions.

\subsection{Intensity Augmentation}
\label{sec:synth}

In addition to warping existing images, \texttt{BrainAug} includes image synthesis from deformed label maps using a combination of standard intensity augmentations, including spatial bias fields, resolution variations, and label-conditioned Gaussian intensity sampling~\cite{billot2023synthseg,iglesias2023synthsr}. This enables the generation of training samples with independently varying anatomy and appearance, reducing reliance on modality-specific intensities. Further details and parameter values are provided in the supplementary material.

\begin{figure*}[htbp]
    \centering
    \includegraphics[width=\linewidth]{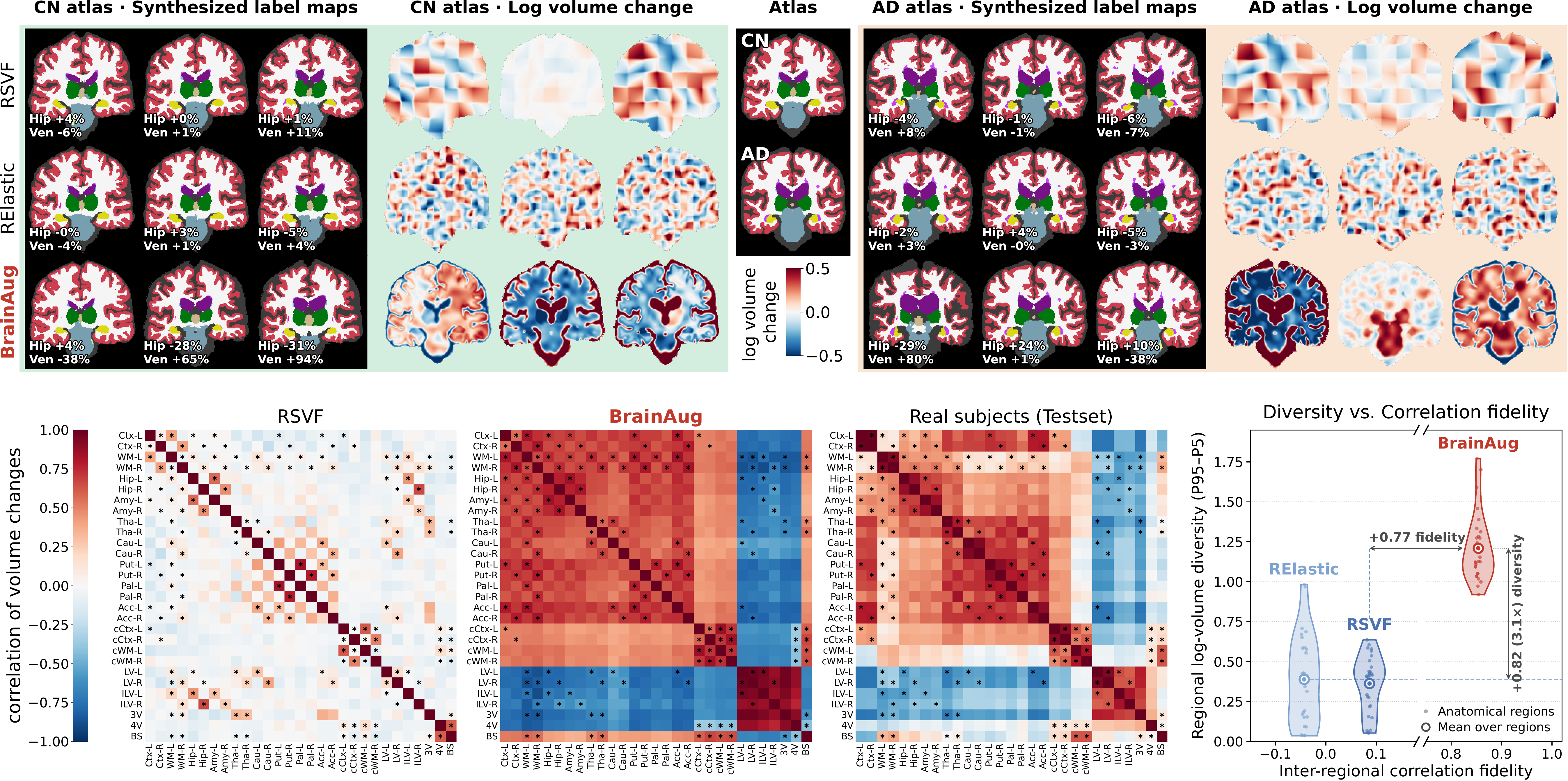}
    \caption{ \textbf{Top:} Synthesized anatomies and log-volume changes from CN and AD atlases; \texttt{BrainAug} yields substantial yet spatially coordinated variations. \textbf{Bottom:} Inter-regional volume correlations (left; $\ast$: structures adjacent in the atlas) and regional log-volume diversity (P95--P5) vs.\ correlation fidelity (right; Pearson $r$ between real and generated correlation matrices). \texttt{BrainAug} closely matches the correlation patterns observed in real subjects, improving fidelity by $+0.77$ over RSVF and diversity by $3.1\times$ over RElastic. }
    \label{fig:divGen}
\end{figure*}

\section{Experimental Evaluation}
\label{sec:exps}

\noindent \textbf{Datasets.} Our experiments span eight public 3D brain MRI datasets:  ABIDE~\cite{di2014autism}, ADHD200~\cite{adhd2012adhd},
ADNI3~\cite{weiner2017alzheimer}, AIBL~\cite{ellis2009australian}, FreeSurfer~\cite{fischl2002whole}, HCP~\cite{van2012human}, OASIS3~\cite{lamontagne2019oasis}, and PPMI~\cite{marek2011parkinson}. The cohorts span childhood to late adulthood, including healthy controls, individuals with autism, and those with neurodegenerative disorders, across T1w, T2w, and FLAIR scans. All scans are registered to MNI152 space and resampled to $182\times218\times182$ at $1$ mm isotropic resolution, with reference segmentations from FreeSurfer~\cite{fischl2012freesurfer}.

\noindent \textbf{Competing Models.} We compare \texttt{BrainAug} against two widely used anatomy-agnostic non-linear spatial augmentation methods: random diffeomorphic (\underline{RSVF}) and random elastic (\underline{RElastic}) deformations~\cite{billot2023synthseg,hoffmann2021synthmorph,hoffmann2024anatomy}. RSVF integrates upsampled random stationary velocity fields to generate diffeomorphic deformations, whereas RElastic interpolates random control-point displacements to produce smooth deformation fields. All methods are combined with random rotations and translations. We also compare against \underline{BrainStorm}~\cite{zhao2019data}, which learns spatial and appearance transformations from labeled and unlabeled MRIs. For registration, we evaluate two \underline{SynthMorph} variants~\cite{hoffmann2021synthmorph,hoffmann2024anatomy}, with and without affine estimation, trained on image pairs synthesized using anatomy-agnostic deformations. 

\noindent \textbf{Implementation Details.} We evaluate \texttt{BrainAug} on segmentation and registration using well-established 3D U-Net architectures, keeping the network design, training protocols, and image synthesis pipeline consistent across augmentation methods to isolate their impact on performance. For segmentation, networks are trained on image--label pairs synthesized on the fly, optimizing a standard loss combining Dice and cross entropy~\cite{isensee2021nnu}. For registration, we train a 3D U-Net that takes moving and fixed images as two input channels and predicts a SVF, optimized using a soft Dice loss and an $L_2$ smoothness penalty~\cite{balakrishnan2019voxelmorph}. Both networks use Adam optimizer (learning rate: $10^{-4}$) and are trained for $200$K steps with a batch size of $1$. Implementation details are provided in the supplementary material.

\subsection{Diversity and Quality of Generated Samples}

To assess the ability of spatial augmentation methods in generating diverse yet anatomically plausible variations from a single atlas, we use cognitively normal (CN) and Alzheimer's disease (AD) reference atlases from OASIS3. We evaluate generated samples from three perspectives: (i)~anatomical diversity and plausibility through label maps and log-Jacobian determinants; (ii)~regional volume dependencies, examining coordinated expansion and contraction against real anatomical relationships; and (iii)~diversity--fidelity analysis, quantifying regional volume variation and agreement with real inter-regional correlations.

\begin{figure*}[htbp]
    \centering
    \includegraphics[width=\linewidth]{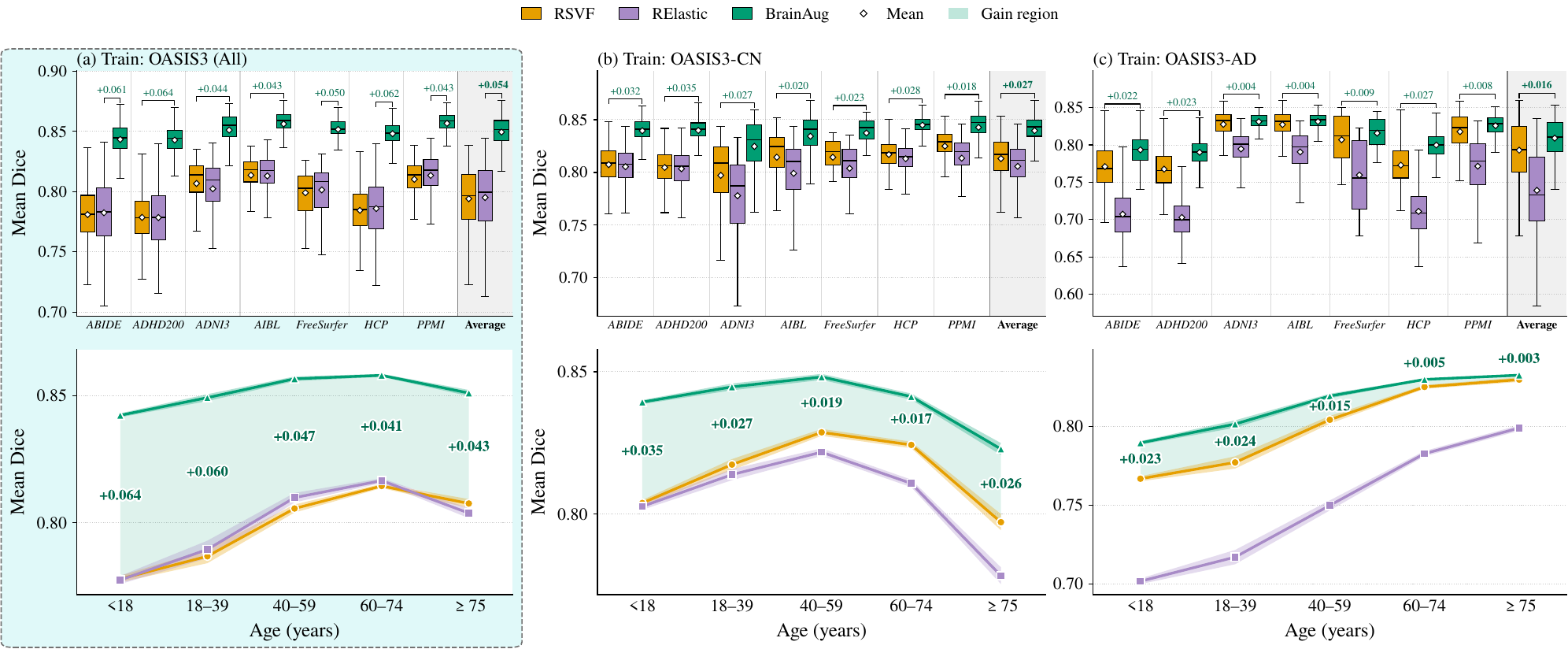}
    \caption{\textbf{Segmentation generalization across seven independent datasets.} \textbf{Top:} Mean Dice distributions of models trained using reference templates derived from the complete OASIS3 cohort (\textcolor[HTML]{19969A}{\textbf{All}}), CN, and AD subgroups. \textbf{Bottom:} Age-stratified performance, showing larger gains with \texttt{BrainAug} among younger subjects, particularly with AD-derived templates.}
    \label{fig:diceGroup}
\end{figure*}

As shown in Fig.~\ref{fig:divGen} (top), RSVF and RElastic approaches introduce substantial local variability, but their deformation patterns are not guided by the underlying anatomy, producing modest regional volume changes. In contrast, \texttt{BrainAug} generates substantial yet spatially coherent volume changes, including coordinated hippocampal shrinkage and ventricular enlargement while preserving anatomical organization. Similar patterns across CN and AD atlases demonstrate its ability to generate structurally coordinated variations across different reference inputs.

The inter-regional volume correlations in Fig.~\ref{fig:divGen} (bottom) show that \texttt{BrainAug} closely matches correlations observed in real subjects (all test datasets except OASIS3), particularly among neighboring and homologous regions, whereas random deformations show much weaker agreement. Measured by Pearson $r$ between off-diagonal entries of real and generated correlation matrices, \texttt{BrainAug} achieves $r=0.854$, while both random baselines remain near zero ($|r|<0.09$). \texttt{BrainAug} also achieves $3.1\times$ greater regional log-volume diversity than the strongest baseline (mean P95--P5: $1.210$ vs.\ $0.389$), demonstrating increased diversity without sacrificing fidelity.

\begin{figure*}[t]
    \centering
    \includegraphics[width=0.70\textwidth]{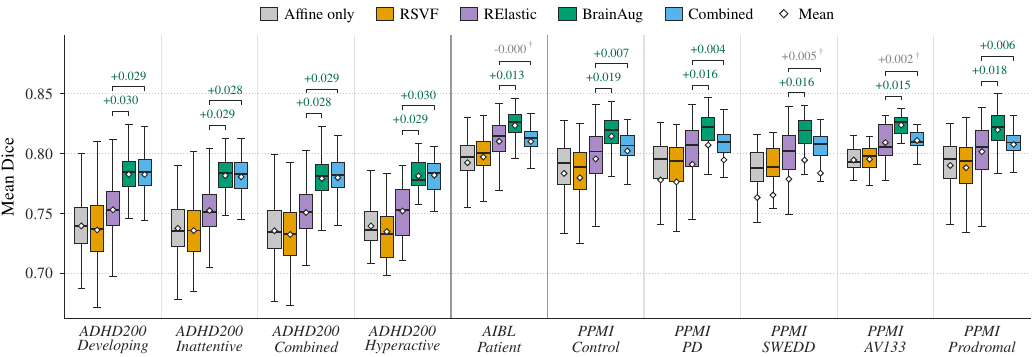}
    \hfill
    \includegraphics[width=0.28\textwidth]{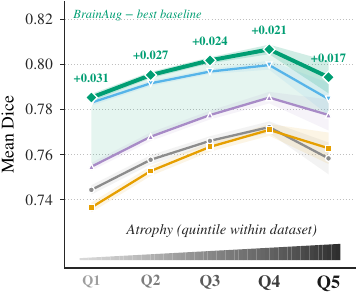}
    \caption{\textbf{Left}: Disease-wise Dice distributions across ten subgroups, comparing five augmentation methods. \textbf{Right:} Segmentation performance across dataset-specific atrophy quintiles, defined using hippocampal, cortical, and ventricular volumes (details in Supplementary). \texttt{BrainAug} outperforms the strongest baseline across all quintiles. Gains are significant (Wilcoxon, Holm-corrected $p<0.05$), except $\dagger$.}
    \label{fig:dice_disease_atrophy}
\end{figure*}

\subsection{Segmentation Generalization Across Diverse Populations}
\label{sec:segGen}

To evaluate whether modeling anatomical dependencies improves the generalizability of segmentation network, we compare \texttt{BrainAug} with the anatomy-agnostic RSVF and RElastic baselines across seven datasets spanning diverse populations and age groups. We train segmentation networks on synthetic images generated from a single OASIS3 reference template under three configurations: the full cohort (All), CN, and AD subpopulations. Following prior work~\cite{balakrishnan2018unsupervised,zhao2019data}, each template is selected as the subject closest to the corresponding population-average anatomy. All methods share the same reference template, intensity augmentation, and training protocol, isolating the contribution of anatomy-aware modeling to segmentation generalization.

Fig.~\ref{fig:diceGroup} (top) shows that \texttt{BrainAug} consistently outperforms RSVF and RElastic across all datasets using the combined OASIS3 template (All), improving average Dice by $5.4$ percentage points (pp) over the strongest baseline. This advantage persists with CN- and AD-specific templates (middle and right), with average gains of $2.7$ and $1.6$ points, respectively. The age-stratified results (bottom) reveal larger improvements among younger subjects, particularly when training from AD-derived anatomy, despite the substantial anatomical differences between the source and target populations. Together, these findings demonstrate that modeling coordinated anatomical variation improves segmentation generalization beyond the reference anatomy, rather than relying solely on random spatial perturbations.

To further compare \texttt{BrainAug} with learned augmentation, we consider BrainStorm~\cite{zhao2019data}, which learns deformation and appearance distributions from unlabeled MRIs as a data-driven alternative. As BrainStorm requires unlabeled MRIs to learn its augmentation model, we train it on the full OASIS3 T1w cohort using the OASIS3 (All) labeled template, whereas \texttt{BrainAug} generates anatomical variations from the same template without additional imaging data. Both methods use identical segmentation training settings. For a fair comparison with BrainStorm, which learns from real MRIs, we apply \texttt{BrainAug}'s generated deformations to the corresponding real T1w template image. Separately, we evaluate cross-contrast generalization by training on images synthesized through intensity augmentation and testing on unseen T2w and FLAIR contrasts.

\begin{table}[htbp]
\centering
\small
\setlength{\tabcolsep}{5pt}
\renewcommand{\arraystretch}{1.08}
\renewcommand{\sd}[1]{\,{\scriptstyle\pm\,#1}}
\caption{Segmentation Dice (mean\,$\pm$\,std) comparing BrainStorm and \texttt{BrainAug}. T2 and FLAIR are unseen contrasts (same ADNI3 subjects as T1). \textbf{Bold}: best; $^{\ast}$: \texttt{BrainAug} (real) better than BrainStorm (paired Wilcoxon, Holm-corrected, $p<10^{-9}$).}
\label{tab:seg}
\begin{adjustbox}{max width=\columnwidth}
\begin{tabular}{@{}ll c >{\columncolor{oursbg}}c >{\columncolor{oursbg}}c@{}}
\toprule
 & & \multicolumn{2}{c}{\textit{Real T1}} & \multicolumn{1}{>{\columncolor{oursbg}}c@{}}{\textit{Synthetic}} \\
\cmidrule(lr){3-4}\cmidrule(l){5-5}
Cont. & Dataset & BrainStorm & \multicolumn{2}{>{\columncolor{oursbg}}c@{}}{\textbf{\texttt{BrainAug}}} \\
\midrule
\multirow{7}{*}{T1}
 & ABIDE     & $0.904\sd{0.083}$ & $\mathbf{0.914}^{\ast}\sd{0.082}$ & $0.836\sd{0.076}$ \\
 & ADHD200   & $0.911\sd{0.031}$ & $\mathbf{0.917}^{\ast}\sd{0.033}$ & $0.842\sd{0.029}$ \\
 & AIBL      & $0.932\sd{0.023}$ & $\mathbf{0.943}^{\ast}\sd{0.022}$ & $0.855\sd{0.025}$ \\
 & FreeSurfer    & $0.924\sd{0.007}$ & $\mathbf{0.936}^{\ast}\sd{0.010}$ & $0.852\sd{0.009}$ \\
 & HCP       & $0.916\sd{0.007}$ & $\mathbf{0.929}^{\ast}\sd{0.008}$ & $0.848\sd{0.009}$ \\
 & PPMI      & $0.920\sd{0.085}$ & $\mathbf{0.934}^{\ast}\sd{0.083}$ & $0.849\sd{0.076}$ \\
 & ADNI3     & $0.932\sd{0.006}$ & $\mathbf{0.944}^{\ast}\sd{0.009}$ & $0.851\sd{0.016}$ \\
\cmidrule(l){2-5}
 & \textit{Average} & $0.917\sd{0.059}$ & $\mathbf{0.928}^{\ast}\sd{0.059}$ & $0.846\sd{0.054}$ \\
\midrule
T2    & ADNI3 & $0.070\sd{0.013}$ & $0.084\sd{0.012}$ & $\mathbf{0.842}\sd{0.017}$ \\
FLAIR & ADNI3 & $0.574\sd{0.016}$ & $0.397\sd{0.025}$ & $\mathbf{0.835}\sd{0.021}$ \\
\bottomrule
\end{tabular}
\end{adjustbox}
\end{table}

As shown in Table~\ref{tab:seg}, \texttt{BrainAug} outperforms BrainStorm across all seven T1w datasets, achieving higher average Dice ($0.928$ vs.\ $0.917$, $p<10^{-9}$) using a single reference anatomy rather than learning augmentation from an unlabeled MRI cohort. Gains are consistent across datasets, demonstrating the effectiveness of anatomy-aware deformation modeling relative to learned augmentation. While both methods trained with real T1w intensities generalize poorly to unseen contrasts, \texttt{BrainAug} with synthetic intensity augmentation maintains consistent performance across all modalities (Dice $0.835$--$0.851$), highlighting the complementary benefit of contrast randomization.

\subsection{Complementarity of Spatial Augmentations}
\label{sec:complement}
To investigate whether conventional random deformations provide complementary benefits to \texttt{BrainAug}, we compare individual augmentation methods with their combination. All methods include random affine transformations, while the Combined setting sequentially composes RSVF, RElastic, and \texttt{BrainAug} deformations. Fig.~\ref{fig:dice_disease_atrophy} (left) shows that \texttt{BrainAug} outperforms the individual baselines across ten subgroups from ADHD200, AIBL, and PPMI. Combining random deformations with \texttt{BrainAug} yields no significant difference on ADHD200 and significantly lowers Dice on AIBL and PPMI (by $1.3$ and $1.5-1.9$ pp; paired Wilcoxon $p < 10^{-100}$), suggesting that random deformations are better replaced than combined with anatomy-aware augmentation. The atrophy-stratified analysis (right) further shows that \texttt{BrainAug} maintains its advantage across all five quintiles, with improvements ranging from $1.7$ to $3.1$ points.

\subsection{Data-Driven Large-Deformation Registration}

Beyond segmentation, we investigate whether \texttt{BrainAug}'s ability to generate large, anatomically coordinated deformations improves registration across substantial morphological differences. Unlike random spatial perturbations, \texttt{BrainAug} models inter-regional dependencies, enabling training pairs with coordinated structural changes, such as ventricular enlargement and hippocampal shrinkage. Following SynthMorph~\cite{hoffmann2021synthmorph}, we train a 3D U-Net using a soft Dice objective on images synthesized from FreeSurfer subjects~\cite{fischl2002whole}. We compare against two pretrained SynthMorph models: SM-Def (deformable)~\cite{hoffmann2021synthmorph} and SM-Joint (affine--deformable)~\cite{hoffmann2024anatomy}. We evaluate $5{,}000$ bidirectional registrations between $50$ ADHD200 subjects (7--26 years) and $50$ AIBL subjects (58--86 years). We report Dice, HD95, ASSD, regional volume mismatch $|\log(V_{\text{moved}}/V_{\text{fixed}})|$, and the fraction of voxels with non-positive Jacobian determinants.

\begin{table}[htbp]
  \centering
  \setlength{\tabcolsep}{3.5pt}
  \renewcommand{\arraystretch}{1.15}
  \caption{Bidirectional inter-subject registration ($5{,}000$ registrations). Mean\,$\pm$\,std over registrations, averaged over $29$ anatomical structures. HD95 and ASSD in mm; Vol.\ err.\ $=|\log(V_{\text{moved}}/V_{\text{fixed}})|$.  \textbf{Bold}: best; \underline{underline}: second best.}
  \resizebox{\columnwidth}{!}{%
  \begin{tabular}{@{}lcccc@{}}
    \toprule
    Method & Dice\,$\uparrow$ & HD95\,$\downarrow$ & ASSD\,$\downarrow$ & Vol.\ err.\,$\downarrow$ \\
    \midrule
    \textcolor{gray}{Affine only}
      & \textcolor{gray}{$0.564{\scriptstyle\,\pm0.059}$}
      & \textcolor{gray}{$4.36{\scriptstyle\,\pm0.71}$}
      & \textcolor{gray}{$1.83{\scriptstyle\,\pm0.33}$}
      & \textcolor{gray}{$0.324{\scriptstyle\,\pm0.100}$} \\
    SM-Def~\cite{hoffmann2021synthmorph}
      & $0.789{\scriptstyle\,\pm0.018}$
      & $2.21{\scriptstyle\,\pm0.28}$
      & $0.81{\scriptstyle\,\pm0.09}$
      & $0.209{\scriptstyle\,\pm0.054}$ \\
    SM-Joint~\cite{hoffmann2024anatomy}
      & \underline{$0.813$}${\scriptstyle\,\pm0.020}$
      & \underline{$2.05$}${\scriptstyle\,\pm0.34}$
      & \underline{$0.76$}${\scriptstyle\,\pm0.10}$
      & \underline{$0.151$}${\scriptstyle\,\pm0.056}$ \\
    \midrule
    \rowcolor{oursrow}
    \textbf{\texttt{BrainAug}}
      & $\mathbf{0.822}{\scriptstyle\,\pm0.014}$
      & $\mathbf{1.94}{\scriptstyle\,\pm0.22}$
      & $\mathbf{0.73}{\scriptstyle\,\pm0.07}$
      & $\mathbf{0.095}{\scriptstyle\,\pm0.025}$ \\
    \bottomrule
  \end{tabular}}
  \label{tab:registration}
\end{table}

\begin{figure}[htbp]
    \centering
    \includegraphics[width=\linewidth, trim = 0cm 0cm 0cm 0cm]{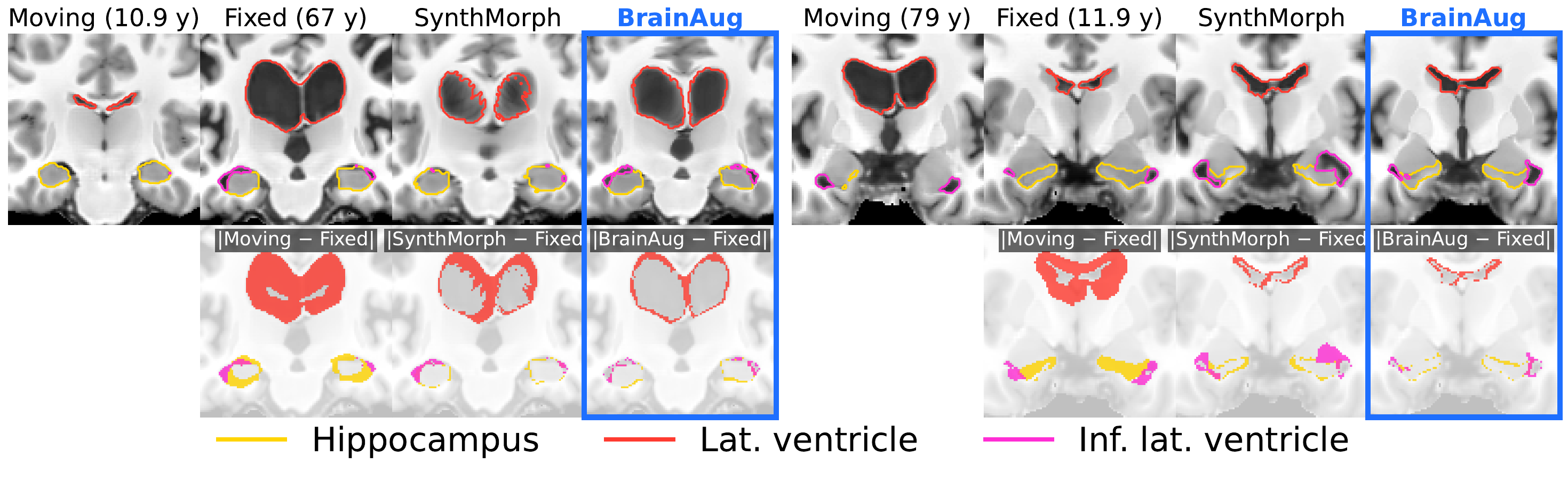}
    \caption{Qualitative comparison of bidirectional registration across large anatomical differences, showing improved hippocampal and ventricular alignment with \texttt{BrainAug}.}
    \label{fig:registration}
\end{figure}

As shown in Table~\ref{tab:registration}, \texttt{BrainAug} outperforms both pretrained SynthMorph models, achieving higher anatomical overlap, lower surface distances, and better regional volume agreement. No non-positive Jacobian determinants were observed, indicating preserved deformation topology in the evaluated registrations. Qualitative results in Fig.~\ref{fig:registration} further illustrate improved hippocampal and ventricular alignment in both registration directions.

\subsection{Ablation Study and Runtime Analysis}

We investigate the contributions of anatomically structured sampling and volume-prescribed deformation generation by ablating (i)~anatomical graph structure, through random permutation of graph connections; (ii)~graph-based correlations $(\beta=0)$, removing graph-induced regional dependencies; (iii)~global latent variation $(\rho=0)$, removing shared brain-wide variation; (iv)~intracranial-volume compensation, disabling tissue--CSF volume balancing; (v)~per-atlas volume calibration; and (vi)~within-region perturbations $(\sigma_\epsilon=0)$, removing local spatial variability. All variants use the OASIS3 AD template with synthetic intensities, are trained under identical settings.

\begin{figure}[htbp]
    \centering
    \includegraphics[width=0.9\linewidth]{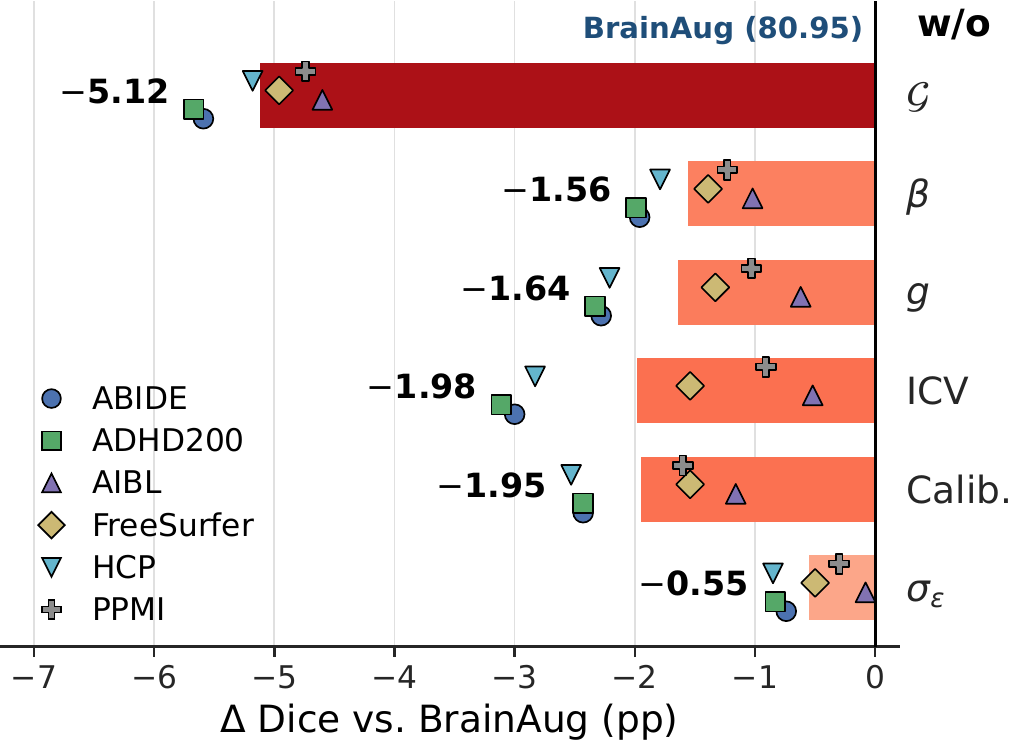}
    \caption{Ablation of our \texttt{BrainAug} model. Bars: change in mean Dice (percentage points) relative to the full \texttt{BrainAug} model (AD template, $80.95$), averaged over six unseen datasets; markers: change on each dataset. Each row removes one component: $\mathcal{G}$: anatomical graph (randomly permuted; Eq.~\ref{eq:graph_g}); $\beta$: graph correlation ($\beta{=}0$; Eq.~\ref{eq:volChange}); $g$: global latent ($\rho{=}0$; Eq.~\ref{eq:corr}); ICV: intracranial-volume conservation (Sec.~\ref{sec:flow}); Calib.: per-atlas volume calibration; $\sigma_\epsilon$: within-region perturbation ($\sigma_\epsilon=0$; Eq.~\ref{eq:epsilon})}
    \label{fig:ablation}
\end{figure}

As shown in Fig.~\ref{fig:ablation}, disrupting the anatomical graph $\mathcal{G}$ causes the largest performance degradation, highlighting the importance of meaningful inter-regional relationships. Removing graph-based correlations ($\beta=0$) or global latent variation ($\rho=0$) also reduces performance, demonstrating the complementary roles of anatomically structured dependencies and brain-wide morphological variation. Disabling intracranial-volume compensation degrades performance, underscoring the importance of balancing tissue contraction with CSF expansion, while removing per-atlas calibration highlights the benefit of adapting volume changes to the reference anatomy. Excluding within-region perturbations ($\sigma_\epsilon = 0$) has the smallest effect, suggesting that coordinated regional volume changes contribute more to performance than local spatial variability.

\noindent \textbf{Runtime.} Our augmentation takes only $79.7{\scriptstyle\,\pm2.1}$\,ms per volume ($182 \times 218 \times 182$) on the GPU, slightly faster than RSVF ($83.4{\scriptstyle\,\pm2.6}$\,ms) and RElastic ($83.2{\scriptstyle\,\pm2.5}$\,ms) under the same integration and warping. The anatomical graph is built only once per atlas ($1.1$\,s) and cached for all subsequent samples; amortized over the $200$K training iterations, this adds only ${\sim}0.006$\,ms per volume (${<}0.01\%$ of the runtime).

\section{Discussion and Conclusion}
\label{sec:con}

In this paper, we introduce \texttt{BrainAug}, an anatomy-aware augmentation framework that generates diverse, coordinated anatomical variations without requiring large imaging cohorts or learned deformation priors. Our experiments demonstrate its effectiveness in improving generalization across brain MRI segmentation and registration tasks.  

Although \texttt{BrainAug} enables substantial anatomical diversity across a broad range of imaging appearances without relying on population imaging data, our segmentation experiments primarily focus a single-template setting to isolate the contribution of anatomically coordinated augmentation. This setting represents an extreme case of limited source diversity, and the relative performance gains are expected to diminish for larger training cohorts encompassing a broader range of anatomical variation. 

Beyond augmentation, future work will investigate the use of \texttt{BrainAug}-generated deformation fields as supervisory signals for learning anatomically informed representations across downstream tasks. With just a few lines of code, \texttt{BrainAug} holds promise to improve the performance of countless brain MRI networks, turning a minimal change in existing training pipelines into an opportunity for far-reaching impact across the field.


{
    \small
    \bibliographystyle{ieeenat_fullname}
    \bibliography{main}
}


\clearpage

\addtocontents{toc}{\protect\setcounter{tocdepth}{2}}

\setcounter{section}{0}
\renewcommand{\thesection}{\Alph{section}}

\counterwithin{figure}{section}
\counterwithin{table}{section}
\counterwithin{equation}{section}

\renewcommand{\thefigure}{\thesection\arabic{figure}}
\renewcommand{\thetable}{\thesection\arabic{table}}
\renewcommand{\theequation}{\thesection\arabic{equation}}

\maketitlesupplementary

\setcounter{tocdepth}{2}
{\small\tableofcontents}
\vspace{1em}

\section{Experimental Dataset}
\label{sec:supp_data}

\subsection{Dataset Details}

\noindent\textbf{Training data.} For segmentation, we use OASIS3~\cite{lamontagne2019oasis} to select reference anatomies from $150$ subjects: $50$ cognitively normal (CN), $50$ with Alzheimer's disease (AD), and $50$ with uncertain diagnoses (\eg, MCI). The single-template experiments use OAS30690 (CN), OAS30043 (AD), and OAS31335 (All), selected as described in Sec.~\ref{sec:atlas_selection}.

\noindent\textbf{Test data.} For segmentation, we evaluate on seven independent test datasets (\cref{tab:supp_data}), comprising $5{,}278$ subjects and $5{,}938$ scans from more than $30$ acquisition sites. ADNI3 provides T1w, T2w, and FLAIR scans of the same $330$ subjects, enabling cross-contrast evaluation while holding subject anatomy fixed.

\noindent\textbf{Labels and evaluation.} Reference segmentations are derived from FreeSurfer
\texttt{aseg} labels~\cite{fischl2012freesurfer},
merged into $25$ structures following
SynthSeg~\cite{billot2023synthseg}: $11$ bilateral regions (cerebral white matter and cortex, lateral ventricle, cerebellar white matter and cortex, thalamus, caudate, putamen, pallidum, hippocampus, and amygdala), and $3$ midline structures (third ventricle, fourth ventricle, and brainstem). All scans are aligned to MNI152 space and resampled to $182\times218\times182$ voxels at $1$ mm isotropic resolution.

\begin{table*}[!t]
\centering
\small
\setlength{\tabcolsep}{4pt}
\begin{tabular}{@{}l r r l c c c p{6.1cm}@{}}
\toprule
Dataset & Subj. & Scans & Contrast & Age (yrs) & Female & Sites & Diagnostic groups ($n$) \\
\midrule
ABIDE~\cite{di2014autism}          & 1{,}099 & 1{,}099 & T1w & -- & -- & 16 & ASD and typical controls$^\dagger$ \\
ADHD200~\cite{adhd2012adhd}        &   961 &   961 & T1w & $12.1\pm3.3$ [7--26] & 39\% & 10 & TDC (576), ADHD-C (209), ADHD-I (137), ADHD-H (13), n/a (26) \\
ADNI3~\cite{weiner2017alzheimer}   &   330 &   990 & T1w, T2w, FLAIR & $78.2\pm7.0$ [61--97] & 48\% & multi & CN (131), EMCI (89), SMC (52), LMCI (37), MCI (19), AD (2) \\
AIBL~\cite{ellis2009australian}    &   668 &   668 & T1w & $72.9\pm6.6$ [55--96] & 56\% & multi & aging cohort: HC, MCI, AD$^\dagger$ \\
FreeSurfer~\cite{fischl2002whole}  &    39 &    39 & T1w & -- & -- & -- & -- \\
HCP~\cite{van2012human}            &   897 &   897 & T1w & 22--36+$^\ddagger$ & 54\%$^\ddagger$ & 1 & healthy young adults \\
PPMI~\cite{marek2011parkinson}     & 1{,}284 & 1{,}284 & T1w & $62.6\pm9.4$ [31--86] & 41\% & multi & PD (626), prodromal (368), HC (206), SWEDD (64), other (20) \\
\midrule
Total                              & 5{,}278 & 5{,}938 & & 7--97 & & $>$30 & \\
\bottomrule
\end{tabular}
\caption{\textbf{Test cohorts.} Age as mean $\pm$ SD [range]; female share among subjects with
recorded sex. $^\ddagger$Age (in 5-year bands) and sex available for 100 of 897 subjects.
ASD: autism spectrum disorder; TDC: typically developing children; ADHD-C/I/H: combined /
inattentive / hyperactive-impulsive subtype; SMC: subjective memory concerns; E/LMCI: early/late
mild cognitive impairment; HC: healthy controls; PD: Parkinson's disease; SWEDD: scans without
evidence of dopaminergic deficit.}
\label{tab:supp_data}
\end{table*}

\subsection{Atlas Selection}
\label{sec:atlas_selection}

\begin{figure*}[t]
    \centering
    \includegraphics[width=\linewidth]{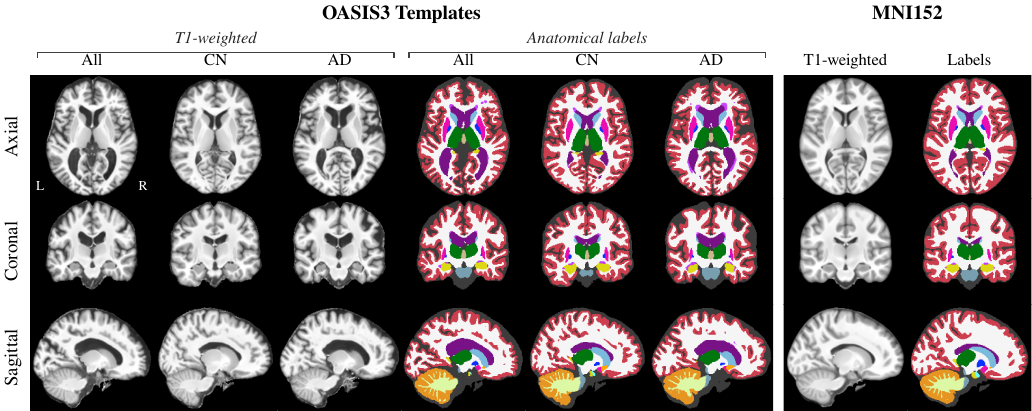}
    \caption{Reference anatomies used in our experiments. T1-weighted images and corresponding anatomical label maps of the OASIS3 All, CN, and AD templates, shown in axial, coronal, and sagittal views. The MNI152 reference is included for visual comparison.}
    \label{fig:allTemp}
\end{figure*}

\begin{figure}[htbp]
    \centering
    \includegraphics[width=\linewidth]{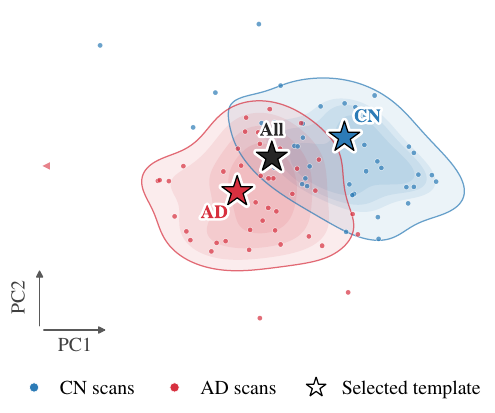}
    \caption{Population-level visualization of the selected OASIS3 templates. PCA projection of standardized regional log-volume fractions from CN (blue) and AD (red) subjects. Stars indicate the All, CN, and AD templates, while shaded contours illustrate the corresponding population distributions.}
    \label{fig:tempPCA}
\end{figure}

We select three OASIS3 reference templates representing the full population (All), cognitively normal (CN), and Alzheimer's disease (AD) subgroups. Following prior work~\cite{balakrishnan2018unsupervised,zhao2019data}, each template is the subject closest to its group-average anatomy in standardized regional log-volume fraction space. Figure~\ref{fig:tempPCA} shows their positions within the CN and AD anatomical distributions, while Figure~\ref{fig:allTemp} compares their T1w images and segmentations with the MNI152 reference.

\section{Implementation Details}
\label{supp:impl}

\subsection{\texttt{BrainAug} Hyperparameters}
\label{supp:hyper}

Table~\ref{tab:supp_hyper} summarizes the \texttt{BrainAug} hyperparameters used across all experiments. The anatomical graph and volume calibration are computed once per reference label map and cached for subsequent augmentations. All experiments use a fixed hyperparameter configuration, selected without tuning on test data.

\begin{table}[h]
\centering\small
\setlength{\tabcolsep}{4pt}
\caption{\texttt{BrainAug} hyperparameters.}
\label{tab:supp_hyper}
\begin{tabular}{@{}l p{0.52\columnwidth} r@{}}
\toprule
Symbol & Description & Value \\
\midrule
\multicolumn{3}{@{}l}{\textit{Anatomical graph (Sec.~\ref{sec:graph})}} \\
$\tau$ & Edge distance threshold & $3\,\mathrm{mm}$ \\
$A^{\mathrm{spat}}_{ij}$ & Spatial edge weight & $1/(1{+}d_{ij})$ \\
$A^{\mathrm{bil}}_{ij}$ & Bilateral edge weight & $0.5\max A^{\mathrm{spat}}$ \\
\midrule
\multicolumn{3}{@{}l}{\textit{Regional volume prior (Sec.~\ref{sec:prior})}} \\
$\beta$ & Heat-kernel diffusion & $1.0$ \\
$\rho$ & Global variance fraction & $0.3$ \\
$\gamma$ & Variation magnitude & $0.25$ \\
$c$ & Log-volume bound & $0.6$ \\
$\alpha$ & Directional pattern strength & $0$ \\
\midrule
\multicolumn{3}{@{}l}{\textit{Deformation generation (Sec.~\ref{sec:flow})}} \\
$\sigma_\epsilon$ & Strength of within-region perturbation & $0.08$ \\
-- & Downsampling factor of $\epsilon$ & $8$ \\
$c_{\mathrm{csf}}$ & CSF divergence clip & $0.7$ \\
$N$ & Calibration samples & $60$ \\
-- & Calibration gain range & $[1,2]$ \\
-- & FFT zero-padding & $16$ vox. \\
-- & Scaling-and-squaring steps & $7$ \\
\bottomrule
\end{tabular}
\end{table}

\subsection{Additional Implementation Details}
\label{supp:impl_details}

\paragraph{Tissue--CSF Compensation.}
To compensate for tissue volume changes, CSF voxels $\mathcal C$ comprise the lateral, inferior lateral, third, fourth, and fifth ventricles and the CSF label. These voxels receive a uniform divergence that offsets the net divergence of tissue voxels $\mathcal T$:
\begin{equation}
f(\mathbf x)=\operatorname{clip}\Big(
-\frac{1}{|\mathcal C|}
\sum_{\mathbf y\in\mathcal T}f(\mathbf y),
-c_{\mathrm{csf}},c_{\mathrm{csf}}
\Big),\qquad \mathbf x\in\mathcal C.
\end{equation}
Without clipping, the divergence sums to zero over the brain, conserving intracranial volume to first order. Clipping limits excessive CSF contraction or expansion, making volume conservation approximate when the bound is reached.

\paragraph{Volume Calibration.}
For each reference label map, we sample $N$ deformations and estimate the relationship between prescribed and realized regional log-volume changes. For each non-CSF region $i$ containing at least $500$ voxels, we regress the realized changes on the prescribed $z_i$ to obtain slope $s_i$. The calibration gain is
\begin{equation}
k_i =
\begin{cases}
\operatorname{clip}(1/s_i,1,2), & s_i>0.2,\\
1, & \text{otherwise}.
\end{cases}
\end{equation}
The calibrated divergence target is then $k_i z_i$. Please note that this calibration is optional for augmentation, as exact agreement between prescribed and realized regional volume changes is not required to generate anatomically diverse samples.

\paragraph{Within-Region Perturbation.}
The perturbation $\epsilon$ in Eq.~\ref{eq:epsilon} introduces smooth spatial variation within anatomical regions. We sample i.i.d.\ Gaussian noise on a grid downsampled by a factor of $8$ and upsample it using cubic interpolation to obtain $\tilde n$. The resulting field is normalized by its spatial standard deviation and restricted to the brain:
\begin{equation}
\epsilon(\mathbf x)=
\sigma_\epsilon
\frac{\tilde n(\mathbf x)}
{\operatorname{std}(\tilde n)}
M(\mathbf x),
\end{equation}
where $M(\mathbf x)\in\{0,1\}$ denotes the brain mask.


\subsection{Mathematical Properties}
\label{supp:math}

\paragraph{Graph Covariance.}
Since the anatomical graph has symmetric, nonnegative edge weights, its Laplacian $L=D-A$ is positive semidefinite. Thus, the heat kernel $K_\beta=\exp(-\beta L)$ is positive definite, and its normalized form $\Sigma_{\mathcal G}=D_K^{-1/2}K_\beta D_K^{-1/2}$, with $D_K=\operatorname{diag}(K_\beta)$, is a valid correlation matrix with unit diagonal.

\paragraph{Divergence and Volume Changes.}
\texttt{BrainAug} constructs a stationary velocity field $\mathbf v=\nabla\psi$ by solving the Poisson equation $\Delta\psi=f$, subject to the solver's compatibility conditions. Let $\varphi_t$ denote its forward flow, satisfying $\partial_t\varphi_t=\mathbf v\circ\varphi_t$ and $\varphi_0=\mathrm{id}$. By Jacobi's formula,
\begin{equation}
\frac{d}{dt}\log\left(\det D\varphi_t(\mathbf{x})\right)
=(\nabla\cdot\mathbf v)(\varphi_t(\mathbf{x})).
\end{equation}
Integrating over $t\in[0,1]$ yields
\begin{equation}
\log\left(\det D\varphi_1(\mathbf{x})\right)
=\int_0^1(\nabla\cdot\mathbf v)(\varphi_t(\mathbf{x}))\,dt
\approx f(\mathbf{x}).
\end{equation}
The approximation holds when the Poisson solution closely reproduces the prescribed divergence and $f$ varies slowly along flow trajectories. Thus, prescribing regional divergence provides approximate control over local log-volume changes, while integrating a sufficiently smooth stationary velocity field yields a diffeomorphic transformation.

\paragraph{Diffeomorphic Property.}
For a sufficiently smooth stationary velocity field with a well-defined flow, the Jacobian determinant satisfies
\begin{equation}
\det\left(D\varphi_t(\mathbf{x})\right)
=\exp\left(
\int_0^t(\nabla\cdot\mathbf v)(\varphi_s(\mathbf{x}))\,ds
\right)>0.
\label{eq:supp_positive_jac}
\end{equation}
Thus, the continuous flow preserves orientation and defines a diffeomorphism, with inverse $\varphi_1^{-1}=\exp(-\mathbf v)$ providing the pullback transformation used for image resampling.


\subsection{Training Sample Generation}
\label{supp:appearance}

\begin{algorithm}[htbp]
\caption{\texttt{BrainAug}: generating one training sample}
\label{alg:sample}
\small
\begin{tcolorbox}[algblock=cGraphA, title={Precomputed once per label map $S$}]
\begin{algorithmic}[1]
\State Build graph $\mathcal G$ and covariance $\Sigma_{\mathcal G}$
    \hfill {\color{gray} Sec.~\ref{sec:graph},~\ref{sec:prior}}
\State Estimate calibration factors $\mathbf k$
    \hfill {\color{gray}Sec. \ref{supp:impl_details}}
\algstore{BrainAug}
\end{algorithmic}
\end{tcolorbox}

\begin{tcolorbox}[algblock=cPriorA, title={Sample regional volume changes}]
\begin{algorithmic}[1]
\algrestore{BrainAug}
\State $\mathbf r\sim\mathcal N(\mathbf 0,\Sigma_{\mathcal G}),\;\;
       g\sim\mathcal N(0,1)$
\State $\mathbf u\gets\sqrt{\rho}\,g\mathbf 1+
       \sqrt{1-\rho}\,\mathbf r$
\State $\omega\sim\mathcal U(\Omega),\;\;
       \tilde{\mathbf u}\gets T_\omega(\mathbf u)$
    \hfill {\color{gray}bilateral mode}
\State $\mathbf z\gets\mathbf k\odot
       c\tanh(\gamma\tilde{\mathbf u}/c)$
    \hfill {\color{gray}calibrated}
\algstore{BrainAug}
\end{algorithmic}
\end{tcolorbox}

\begin{tcolorbox}[algblock=cFlowA, title={Generate the deformation}]
\begin{algorithmic}[1]
\algrestore{BrainAug}
\State $f\gets z_S+\epsilon$ on tissue $\mathcal T$
\State $f\gets\operatorname{clip}\left(
       -\frac{1}{|\mathcal C|}\sum_{\mathcal T}f,
       -c_{\mathrm{csf}},c_{\mathrm{csf}}
       \right)$ on CSF $\mathcal C$
\State $\psi\gets\Delta^{-1}f$ (FFT), \;\;
       $\mathbf v\gets\nabla\psi$
\State $\phi\gets\exp(-\mathbf v)$
    \hfill {\color{gray}scaling and squaring}
\algstore{BrainAug}
\end{algorithmic}
\end{tcolorbox}

\begin{tcolorbox}[algblock=cAppA, title={Synthesize the image}]
\begin{algorithmic}[1]
\algrestore{BrainAug}
\State $S_\phi\gets S\circ\phi$
\State $I\gets\mathrm{GMM}(S_\phi)$ with bias field and gamma
    \hfill {\color{gray}synthetic setting}
\State Apply random rotation to $(I,S_\phi)$
\State \Return a $96^3$ patch of $(I,S_\phi)$
\end{algorithmic}
\end{tcolorbox}
\end{algorithm}

Training samples are generated on the fly following Alg.~\ref{alg:sample}. For each sampled label map, the anatomical graph, covariance, and calibration factors are retrieved from the cache. Regional volume changes, deformations, and image appearances are sampled anew for each iteration. The deformation and rigid transform are composed before resampling.

\paragraph{Appearance.}
Following SynthSeg~\cite{billot2023synthseg}, each label $k$ receives an intensity mean $\mu_k\sim\mathcal U(0,255)$ and standard deviation $\sigma_k\sim\mathcal U(0,35)$, with voxel intensities sampled as $I(\mathbf x)\sim \mathcal N(\mu_{S_\phi(\mathbf x)}, \sigma^2_{S_\phi(\mathbf x)})$. A smooth multiplicative bias field $\exp(b)$ is applied, where $b$ is obtained by trilinearly upsampling a $4^3$ grid of $\mathcal N(0,\sigma_B^2)$ values, with $\sigma_B\sim\mathcal U(0,0.5)$. The image is then min--max normalized and gamma-transformed as $I\gets I^{\exp(\eta)}$, where $\eta\sim\mathcal N(0,0.4^2)$.

For augmentation of real T1w images, the original image
is resampled using the same spatial transform, followed
by a bias field with $\sigma_B\sim\mathcal U(0,0.3)$,
99.5th-percentile normalization, gamma augmentation
with $\eta\sim\mathcal N(0,0.3^2)$, and Gaussian noise
with standard deviation sampled from $\mathcal U(0,0.03)$.

\paragraph{Patch Sampling.}
\label{supp:sampling}
A $96^3$ patch is centered on a randomly selected
foreground voxel with probability $0.8$, and at a
uniformly random location otherwise.

\subsection{Baseline Augmentation Models}
\label{supp:baselines}

All methods, including \texttt{BrainAug} and the baselines, follow the same training pipeline (Alg.~\ref{alg:sample}), sharing rigid augmentation, appearance synthesis, and patch sampling (Sec.~\ref{supp:sampling}). Rotations and translations are sampled independently along each axis from $\mathcal{U}(-20^\circ,20^\circ)$ and $\mathcal{U}(-10,10)$\,mm, respectively, without scaling or shear. Let $\phi$ denote the nonrigid pull-back deformation generated by \texttt{BrainAug} or the corresponding baseline, and $\mathcal{R}$ the shared rigid pull-back mapping. The complete transformation is $\Phi=\phi\circ\mathcal{R}$, and the input label map $S$ is resampled once as $S'=S\circ\Phi$, before appearance synthesis. Here, $\circ$ denotes function composition, and $\mathcal U(a,b)$ denotes the uniform distribution over $[a,b]$. Thus, the methods differ only in how $\phi$ is generated.

\paragraph{Random SVF.}
Random stationary velocity fields (SVFs) generate smooth, anatomy-agnostic deformations by sampling Gaussian velocity vectors on a coarse grid and integrating the interpolated field~\cite{billot2023synthseg}. The velocity magnitude is controlled by a randomly sampled standard deviation, yielding
\begin{equation}
\begin{aligned}
s &\sim \mathcal U(0,s_{\max}), \quad
\mathbf v_0(\mathbf p)\overset{\mathrm{i.i.d.}}{\sim}
\mathcal N(\mathbf 0,s^2\mathbf I),\\
\phi_{\mathrm{SVF}} &= \exp\big(\mathrm U_{\uparrow}(\mathbf v_0)\big),\\
\Phi_{\mathrm{SVF}} &= \phi_{\mathrm{SVF}}\circ\mathcal{R},\quad
S'=S\circ\Phi_{\mathrm{SVF}}.
\end{aligned}
\end{equation}
Here, $\mathbf p$ indexes a $10^3$ control-point grid, $\mathbf v_0$ denotes the sampled velocity vectors, and $\mathcal N(\mathbf 0,s^2\mathbf I)$ is an isotropic Gaussian distribution with standard deviation $s$ per component, where $\mathbf I$ is the identity matrix. The operator $\mathrm U_{\uparrow}$ denotes trilinear upsampling of the coarse vector field to full image resolution, and $s_{\max}=3$\,mm. The exponential denotes stationary velocity integration using $7$ scaling-and-squaring steps, as in \texttt{BrainAug}.

\paragraph{Random Elastic.}
Unlike random SVF, random elastic augmentation generates anatomy-agnostic deformations by directly interpolating randomly sampled displacement vectors, without velocity integration. The displacement field is constructed as
\begin{equation}
\begin{aligned}
\mathbf d_0(\mathbf p) &\overset{\mathrm{i.i.d.}}{\sim}
\mathcal U([-d_{\max},d_{\max}]^3),\\
\phi_{\mathrm{ela}}(\mathbf x)
&=\mathbf x+\mathrm U_{\uparrow}(\mathbf d_0)(\mathbf x),\\
\Phi_{\mathrm{ela}}&=\phi_{\mathrm{ela}}\circ\mathcal{R},\quad
S'=S\circ\Phi_{\mathrm{ela}}.
\end{aligned}
\end{equation}
Here, $\mathbf d_0$ denotes displacement vectors sampled independently on an $8^3$ control-point grid, with each component uniformly distributed over $[-d_{\max},d_{\max}]$, where $d_{\max}=4$\,mm. The coordinate $\mathbf x$ denotes a spatial position in the image domain. Unlike SVF-based deformations, the resulting transformation is not guaranteed to be diffeomorphic.

\paragraph{Combined.}
We additionally evaluate whether combining \texttt{BrainAug} with anatomy-agnostic augmentations improves segmentation generalization beyond using each method independently. To this end, \texttt{BrainAug}, random SVF, and random elastic deformations are sampled independently using their respective settings and sequentially composed as
\begin{equation}
\begin{aligned}
\phi_{\mathrm{comb}}
&=\phi_{\mathrm{ela}}\circ\phi_{\mathrm{SVF}}
\circ\phi_{\mathrm{\texttt{BrainAug}}},\\
\Phi_{\mathrm{comb}}
&=\phi_{\mathrm{comb}}\circ\mathcal{R},\quad
S'=S\circ\Phi_{\mathrm{comb}}.
\end{aligned}
\end{equation}
Here, $\phi_{\mathrm{\texttt{BrainAug}}}$ denotes the anatomy-aware deformation generated by \texttt{BrainAug}, and $\phi_{\mathrm{comb}}$ the composition of all three nonrigid deformations. This setting evaluates whether additional deformation diversity complements \texttt{BrainAug}'s anatomically coordinated transformations or instead disrupts their structural dependencies. All transformations are composed before a single resampling, with the same appearance synthesis and training protocol as the other methods.

\subsection{Segmentation Network and Training}
\label{supp:training}

\paragraph{Network.}
We use a 3D U-Net~\cite{ronneberger2015u} with five resolution levels and $32$ feature channels at the first level, increasing to $64$, $128$, $256$, and $256$ at subsequent levels. Each level consists of two $3\times3\times3$ convolutions, each followed by group normalization ($8$ groups) and ReLU activation. Downsampling uses max pooling, while upsampling uses transposed convolutions with skip connections. A final $1\times1\times1$ convolution produces logits for $K=26$ classes, including background, which are converted to probabilities using softmax. The same network is used for all augmentation methods. We optimize the sum of soft Dice and cross-entropy losses with equal weights, including background in both terms.

\paragraph{Loss.}
Let $p_k(\mathbf x)$ be the predicted probability of class $k$ at voxel $\mathbf x$ of a patch $\Omega_p$, and $y_k(\mathbf x)\in\{0,1\}$ the one-hot reference. We minimize $\mathcal L=\mathcal L_{\mathrm{Dice}}+\mathcal L_{\mathrm{CE}}$, with
\begin{align}
\mathcal L_{\mathrm{Dice}} &= 1-\frac{1}{K}\sum_{k=1}^{K}
\frac{2\sum_{\mathbf x} p_k(\mathbf x)\,y_k(\mathbf x)+\delta}
{\sum_{\mathbf x} p_k(\mathbf x)+\sum_{\mathbf x} y_k(\mathbf x)+\delta},\\
\mathcal L_{\mathrm{CE}} &= -\frac{1}{|\Omega_p|}\sum_{\mathbf x\in\Omega_p}\sum_{k=1}^{K} y_k(\mathbf x)\log p_k(\mathbf x),
\end{align}
where $\delta$ is a small constant for numerical stability.

\paragraph{Optimization.}
Training samples are generated on the fly following Alg.~\ref{alg:sample}. We use Adam with learning rate $10^{-4}$ and batch size $1$, and train for 200K iterations. All models are trained for 200K iterations using an identical training budget, and the final checkpoint is used for evaluation without validation-based model selection.

\paragraph{Labels.}
We retain $25$ anatomical structures ($11$ bilateral pairs and $3$ unilateral structures), mapping all other labels to background. The retained labels are remapped to a fixed vocabulary of $K=26$ classes, including background, shared across all experiments.

\paragraph{Inference.}
Test images are skull-stripped using brain masks derived from their reference segmentations and normalized to $[0,1]$ using the 99.5th percentile of nonzero intensities. Full-volume inference is performed in a single forward pass. Dice is computed per anatomical class and averaged over foreground classes present in the reference segmentation.

\subsection{Evaluation Protocols}

\paragraph{Atrophy-Stratified Evaluation.}
To evaluate segmentation performance across different degrees of anatomical atrophy, we define a composite score based on hippocampal, cortical, and ventricular volumes obtained from FreeSurfer segmentations. For each subject, the three volumes are normalized by intracranial volume (ICV), and the atrophy score is computed as
\begin{equation}
    a = \frac{1}{3}\left[
    -\mathcal{Z}\left(\frac{V_{\mathrm{hip}}}{V_{\mathrm{ICV}}}\right)
    -\mathcal{Z}\left(\frac{V_{\mathrm{ctx}}}{V_{\mathrm{ICV}}}\right)
    +\mathcal{Z}\left(\frac{V_{\mathrm{vent}}}{V_{\mathrm{ICV}}}\right)
    \right],
\end{equation}
where $V_{\mathrm{hip}}$, $V_{\mathrm{ctx}}$, and $V_{\mathrm{vent}}$ denote hippocampal, cortical, and ventricular volumes, respectively, and $\mathcal{Z}(\cdot)$ denotes z-score standardization across subjects. Higher scores therefore correspond to smaller hippocampal and cortical volumes and larger ventricles. Subjects are ranked by $a$ separately within each dataset and partitioned into five equal-sized quintiles (Q1--Q5). Segmentation performance is then aggregated across the six evaluation datasets for each corresponding quintile.

\section{Extended Evaluation of \texttt{BrainAug}-Synthesized Images}

\subsection{Qualitative Analysis of \texttt{BrainAug}-Generated Anatomical Variations}
\begin{figure*}[htbp]
    \centering
    \includegraphics[width=\linewidth,height=0.85\textheight,keepaspectratio]{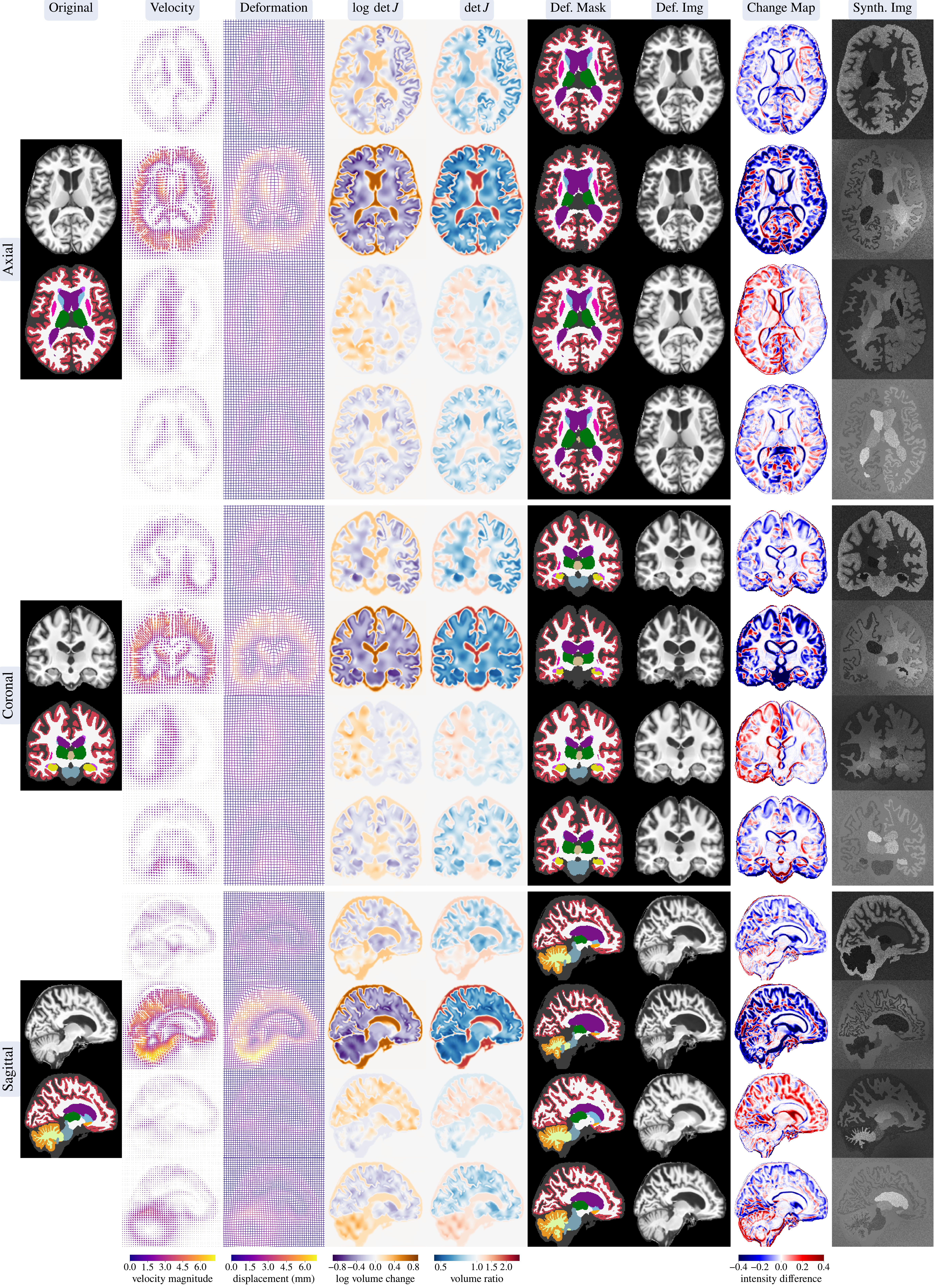}
    \caption{\textbf{Qualitative visualization of \texttt{BrainAug}-generated anatomical variations.} Four independently sampled transformations are shown across axial, coronal, and sagittal views of a single reference anatomy. From left to right: original MRI and segmentation, velocity fields, deformation grids, log-Jacobian determinants ($\log\det J$), Jacobian determinants ($\det J$), deformed segmentations, deformed MRIs, intensity difference maps, and synthesized MRIs. The Jacobian maps illustrate spatially varying expansion and contraction, while the deformed images and segmentations demonstrate anatomical consistency under substantial structural changes.}
    \label{fig:synthImgs}
\end{figure*}

Figure~\ref{fig:synthImgs} illustrates \texttt{BrainAug}'s deformation generation process across axial, coronal, and sagittal views, highlighting three key properties.
\textbf{(i) Spatial coherence:} The velocity fields and deformation grids exhibit smooth, continuous transformations across anatomical structures, reflecting the use of a shared deformation field.
\textbf{(ii) Regional volume modulation:} We visualize local volume changes using the Jacobian determinant $\det J_\phi$, which measures local volume scaling, and its logarithm $\log\det J_\phi$, where positive and negative values indicate local expansion and contraction, respectively. The log-Jacobian maps illustrate spatially heterogeneous volume changes over a displayed range of approximately $[-0.8,0.8]$, while the determinant maps show corresponding local volume scaling over $[0.5,2.0]$. Positive Jacobian determinants indicate locally orientation-preserving deformations.
\textbf{(iii) Anatomical consistency:} The deformed segmentations and corresponding MRIs maintain spatial correspondence despite pronounced changes in ventricular geometry and surrounding structures. The synthesized images further demonstrate the integration of anatomical deformation with appearance augmentation.

\subsection{Bilateral Sampling Modes}

To control hemispheric variation, we consider three bilateral sampling modes: symmetric, asymmetric, and mixed. For each homologous pair $(L,R)$, the graph-correlated latent variables from Eq.~\ref{eq:corr} are transformed as
\begin{equation}
\tilde{u}_L=u_L,\qquad
\tilde{u}_R=b\,u_L+(1-b)\,u_R,
\end{equation}
where the sampling distribution of $b$ determines the bilateral mode:
\begin{equation}
b\sim
\begin{cases}
\delta_1, & \text{symmetric},\\
\delta_0, & \text{asymmetric},\\
\mathrm{Bernoulli}(0.5), & \text{mixed}.
\end{cases}
\end{equation}
Here, $\delta_0$ and $\delta_1$ denote point masses at $0$ and $1$. Symmetric sampling assigns identical latent variations to homologous regions, whereas asymmetric sampling preserves their original graph-correlated variations. Mixed sampling independently selects between symmetric ($b=1$) and asymmetric ($b=0$) modes for each homologous pair with equal probability. 

\begin{figure*}[htbp]
    \centering
    \includegraphics[width=\linewidth,height=0.9\textheight,keepaspectratio]{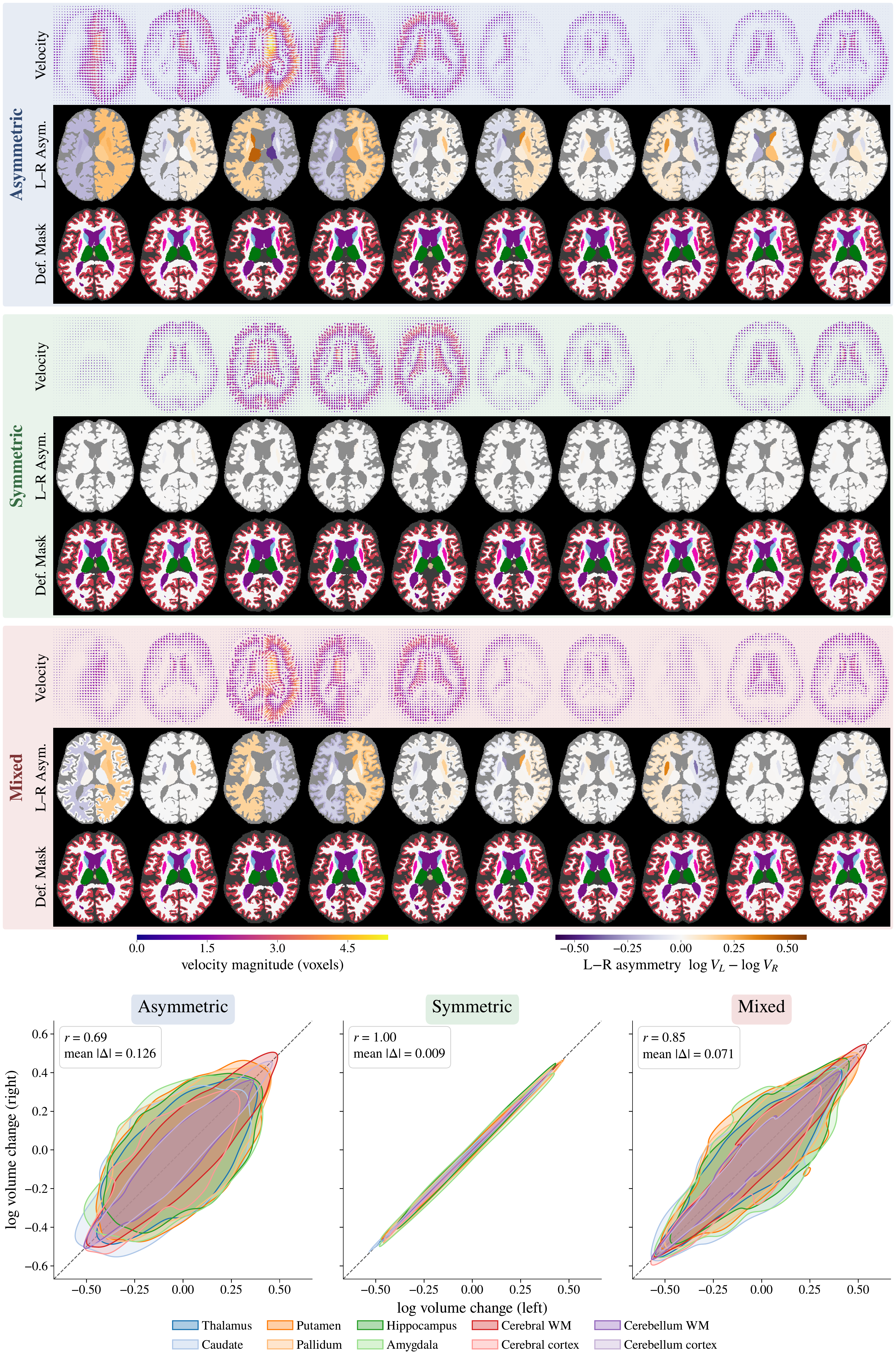}
    \caption{\textbf{Effect of bilateral sampling modes on anatomical variation.} Top: Velocity fields, left--right log-volume asymmetry maps, and deformed segmentations across asymmetric, symmetric, and mixed sampling modes. Bottom: Corresponding bilateral regional log-volume changes, with Pearson correlation ($r$) and mean absolute difference ($|\Delta|$), demonstrating distinct levels of hemispheric asymmetry.}

    \label{fig:velMode}
\end{figure*}

We evaluate how the three bilateral sampling modes control hemispheric anatomical variation. Figure~\ref{fig:velMode} (top) compares the resulting velocity fields, left--right volume asymmetry, and deformed segmentations, showing distinct patterns of bilateral variation. The bottom panels quantify these differences: symmetric sampling achieves nearly identical bilateral changes ($r=1.00$, mean $|\Delta|=0.009$), whereas asymmetric sampling permits greater variation ($r=0.69$, $0.126$). Mixed sampling provides an intermediate behavior ($r=0.85$, $0.071$), demonstrating controllable levels of hemispheric asymmetry while retaining graph-correlated anatomical variation.

\section{Anatomical Segmentations}

\subsection{Comparison with BrainStorm on real T1w images.} 

\begin{figure*}[htbp]
  \centering
  \includegraphics[width=\linewidth]{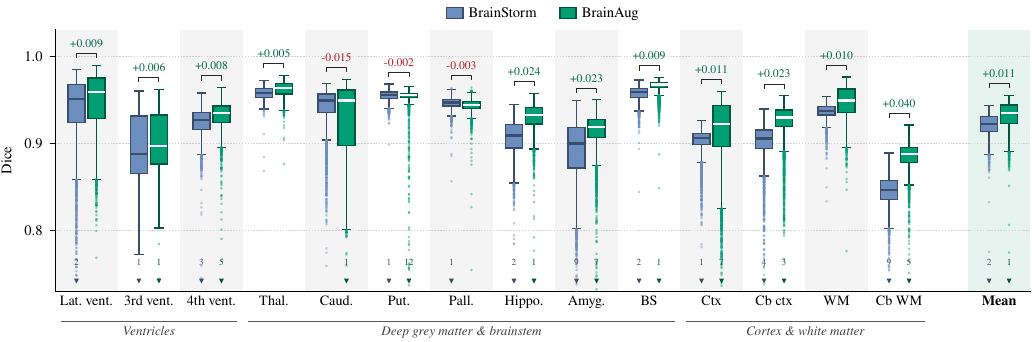}
  \caption{\textbf{Structure-level comparison with BrainStorm (real T1w training).}   Dice per structure (left and right averaged) on the $7$ T1w test datasets.  Both models are trained for 200K iterations using the same OASIS3 (All) reference template. For a fair comparison with BrainStorm, BrainAug applies sampled deformations directly to the real T1w template image, without intensity synthesis. Brackets show the difference in mean Dice   between \texttt{BrainAug} and BrainStorm (green: significant gain; red:   significant loss; paired Wilcoxon, Holm-corrected, $p<0.05$). Numbers with   $\blacktriangledown$ count subjects below the plotted range.   Lat.\ vent.: lateral ventricle; Thal.: thalamus; Caud.: caudate; Put.: putamen;   Pall.: pallidum; Hippo.: hippocampus; Amyg.: amygdala; BS: brainstem;   Ctx: cerebral cortex; Cb: cerebellar; WM: white matter.}
  \label{fig:supp_bstorm_box}
\end{figure*}

\begin{figure*}[t]
  \centering
  \includegraphics[width=\linewidth,height=0.92\textheight,keepaspectratio]{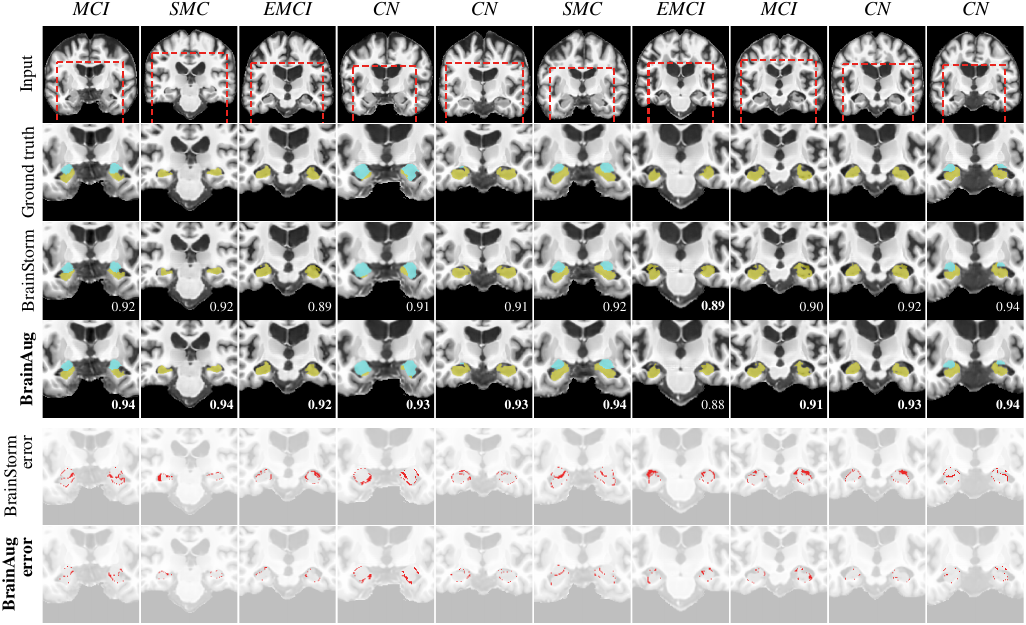}
  \caption{\textbf{Qualitative comparison with BrainStorm on T1w (ADNI3).}   Each column shows one subject (diagnostic group on top). \emph{Input:} coronal   slice with the zoomed region (red). \emph{Ground truth, BrainStorm,   \texttt{BrainAug}:} hippocampus and amygdala of both hemispheres, with their   3-D Dice (best in bold). \emph{Error:} voxels where each prediction disagrees   with the ground truth. Regions are zoomed where the methods differ most.}
  \label{fig:supp_bstorm_t1}
\end{figure*}

Figure~\ref{fig:supp_bstorm_box} extends the real-T1w comparison of Table~\ref{tab:seg} to individual anatomical structures. Across $7$ T1w test datasets, \texttt{BrainAug} improves mean Dice by $1.1$\,pp and outperforms BrainStorm on $11$ of $14$ structures. The largest gains occur in cerebellar white matter ($+4.0$\,pp), hippocampus ($+2.4$\,pp), amygdala ($+2.3$\,pp), and cerebellar cortex ($+2.3$\,pp). BrainStorm performs better on the caudate ($-1.5$\,pp), pallidum ($-0.3$\,pp), and putamen ($-0.2$\,pp).

Figure~\ref{fig:supp_bstorm_t1} provides a qualitative comparison on $10$ ADNI3 subjects, including $6$ with the largest whole-brain gains and $4$ with comparable performance between methods. For the bilateral hippocampus and amygdala, \texttt{BrainAug} achieves higher Dice in $8$ of $10$ subjects, with one tie and one lower score. The error maps show fewer boundary disagreements for \texttt{BrainAug} in most examples, particularly around the hippocampus and amygdala. Together, these results demonstrate that our anatomy-aware deformation improves structure-level segmentation even when both methods are trained from the same real T1w reference image.

\subsection{Generalization to Unseen Contrasts}
\begin{figure*}[t]
  \centering
  \includegraphics[width=\linewidth]{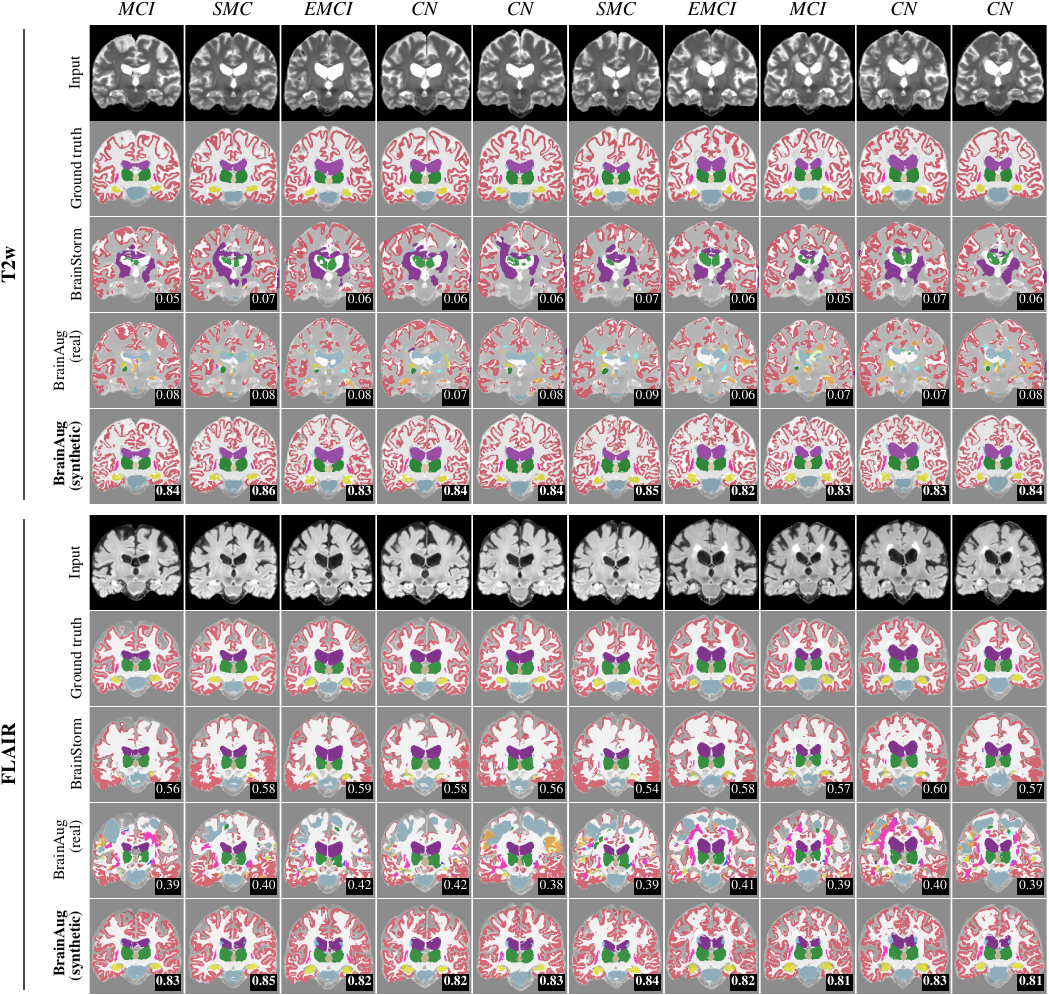}
  \caption{\textbf{Generalization to unseen contrasts (ADNI3).} The same $10$   subjects as in Figure~\ref{fig:supp_bstorm_t1}, imaged with T2w (top) and FLAIR   (bottom). BrainStorm and \texttt{BrainAug} (real) are trained on the real T1w   template image; \texttt{BrainAug} (synthetic) is trained on images synthesized   from the same template. Each panel shows a coronal slice with the whole-volume   mean Dice (best in bold).}
  \label{fig:supp_contrast}
\end{figure*}

Figure~\ref{fig:supp_contrast} evaluates the three models from Table~\ref{tab:seg} on T2w and FLAIR scans of the same $10$ ADNI3 subjects shown in Figure~\ref{fig:supp_bstorm_t1}. Both real-T1w-trained models exhibit substantial segmentation failures on unseen contrasts, particularly on T2w. On FLAIR, BrainStorm retains partial anatomical agreement. In contrast, \texttt{BrainAug} (synthetic) maintains consistent segmentation accuracy across both contrasts (Dice $0.81$--$0.86$). On T2w, BrainStorm incorrectly assigns ventricular labels to substantial portions of the deep grey matter, while \texttt{BrainAug} (real) fails to recover most anatomical structures, yielding Dice scores of $0.05$--$0.09$. On FLAIR, both models produce substantial segmentation errors, particularly in cortical and white-matter regions (BrainStorm: $0.54$--$0.60$; \texttt{BrainAug} (real): $0.38$--$0.42$). In contrast, \texttt{BrainAug} (synthetic) maintains accurate segmentations across both contrasts, achieving Dice scores of $0.82$--$0.86$ on T2w and $0.81$--$0.85$ on FLAIR. These results demonstrate that combining anatomy-aware augmentation with contrast-agnostic image synthesis enables generalization to unseen MRI contrasts without retraining.

\subsection{Segmentation Generalization: Structure-Level Analysis}
\label{supp:structure_all}

We extend the segmentation generalization analysis of Sec.~\ref{sec:segGen} (Figure~\ref{fig:diceGroup}a, OASIS3-All template) to individual anatomical structures. The RSVF, RElastic, and \texttt{BrainAug} networks are trained on images synthesized from the same OASIS3-All reference template and evaluated on $5{,}250$ subjects from $7$ datasets (ABIDE, ADHD200, ADNI3, AIBL, FreeSurfer, HCP, and PPMI). Dice is averaged across hemispheres for each of the $14$ structures. We compare \texttt{BrainAug} against the stronger baseline per structure using paired Wilcoxon signed-rank tests, Holm-corrected across $15$ comparisons, including the overall mean.

\begin{figure*}[t]
  \centering
  \includegraphics[width=\linewidth]{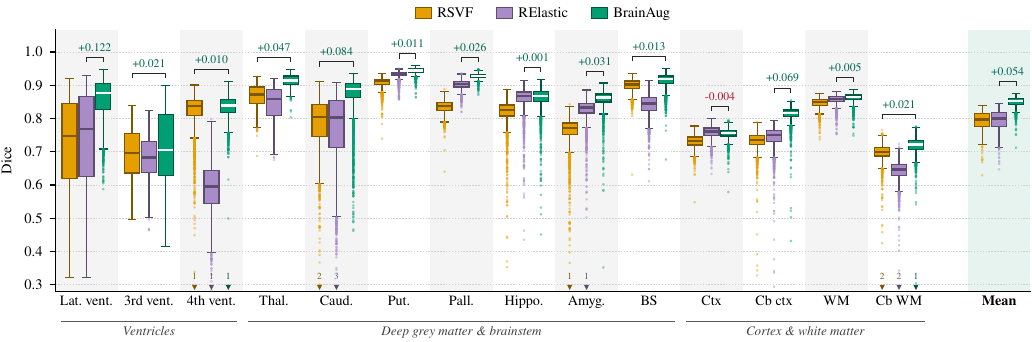}
  \caption{\textbf{Structure-level segmentation accuracy (OASIS3-All template).} Dice per structure (left and right averaged). Brackets indicate the mean Dice difference between \texttt{BrainAug} and the stronger baseline (green: significant gain; red: significant loss; paired Wilcoxon, Holm-corrected, $p<0.05$). Numbers with $\blacktriangledown$ indicate subjects below the plotted range. Lat.\ vent.: lateral ventricle; Thal.: thalamus; Caud.: caudate; Put.: putamen; Pall.: pallidum; Hippo.: hippocampus; Amyg.: amygdala; BS: brainstem; Ctx: cerebral cortex; Cb: cerebellar; WM: white matter.}
  \label{fig:supp_structure_box}
\end{figure*}

Figure~\ref{fig:supp_structure_box} shows that \texttt{BrainAug} improves mean Dice by $5.4$ percentage points (pp) over the stronger baseline, with improvements on $13$ of $14$ structures. The largest improvements occur in the lateral ventricle ($+12.2$\,pp), caudate ($+8.4$\,pp), cerebellar cortex ($+6.9$\,pp), and thalamus ($+4.7$\,pp). Additional gains are observed in the amygdala ($+3.1$\,pp), pallidum ($+2.6$\,pp), and third ventricle ($+2.1$\,pp), while improvements are smaller for the remaining structures ($+0.1$ to $+2.1$\,pp). The cerebral cortex is the only structure with a slight decrease ($-0.4$\,pp). Notably, the lateral ventricle and caudate exhibit substantially reduced lower-tail errors under \texttt{BrainAug}, indicating more consistent segmentation across anatomically diverse subjects. These findings suggest that explicitly modeling coordinated regional volume changes is particularly beneficial for structures exhibiting substantial anatomical variability.

\begin{figure*}[t]
  \centering
  \includegraphics[width=\linewidth,height=0.92\textheight,keepaspectratio]{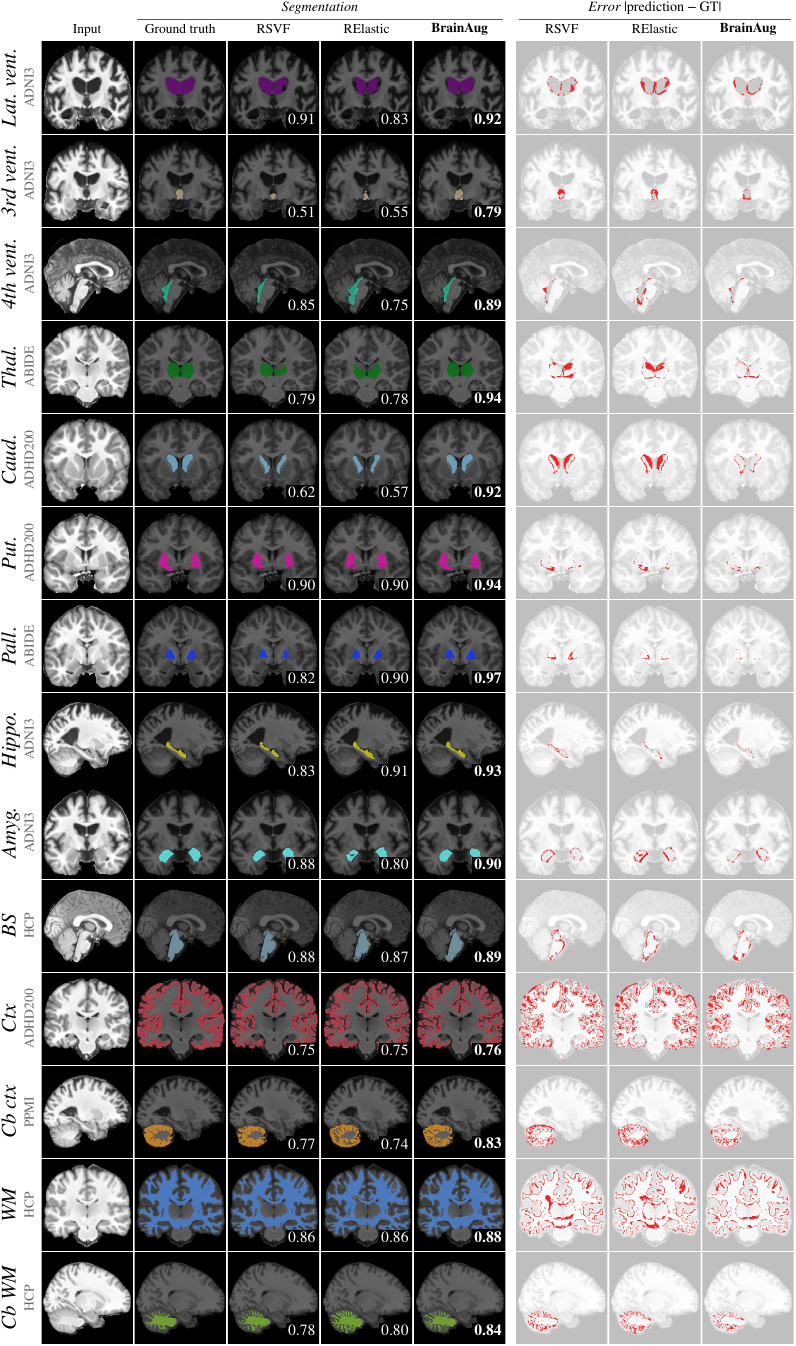}
  \caption{\textbf{Qualitative structure-level segmentation comparison (OASIS3-All template).} Each row presents one anatomical structure from a different test subject (dataset indicated alongside). Ground truth and predictions from RSVF, RElastic, and \texttt{BrainAug} are shown with slice-level Dice (best in bold). Error maps highlight voxels where predictions disagree with the ground truth.}
  \label{fig:supp_structure_qual}
\end{figure*}

For qualitative evaluation, Figure~\ref{fig:supp_structure_qual} presents one subject per structure, selected from cases with substantial \texttt{BrainAug} gains and sufficient anatomical visibility. Subjects with MCI or AD are prioritized for the ventricles, hippocampus, and amygdala. Each structure is shown in its most informative coronal or sagittal view, with slice-level Dice and voxelwise segmentation errors. These examples illustrate characteristic improvements rather than average performance.

Figure~\ref{fig:supp_structure_qual} illustrates these improvements across representative anatomical structures. \texttt{BrainAug} achieves the highest slice-level Dice in all $14$ selected examples, with particularly large gains in the caudate ($0.92$ vs.\ $0.62$), third ventricle ($0.79$ vs.\ $0.55$), and thalamus ($0.94$ vs.\ $0.79$). The corresponding error maps show that anatomy-agnostic baselines frequently miss portions of these structures, whereas \texttt{BrainAug} more closely follows their anatomical boundaries. Differences are smaller for the cerebral cortex ($0.76$ vs.\ $0.75$) and cerebral white matter ($0.88$ vs.\ $0.86$), consistent with the quantitative results.

\subsection{Qualitative Segmentation Across Clinical Subgroups}
\label{supp:per_structure}

\begin{figure*}[!t]
    \centering
    \includegraphics[width=\linewidth]{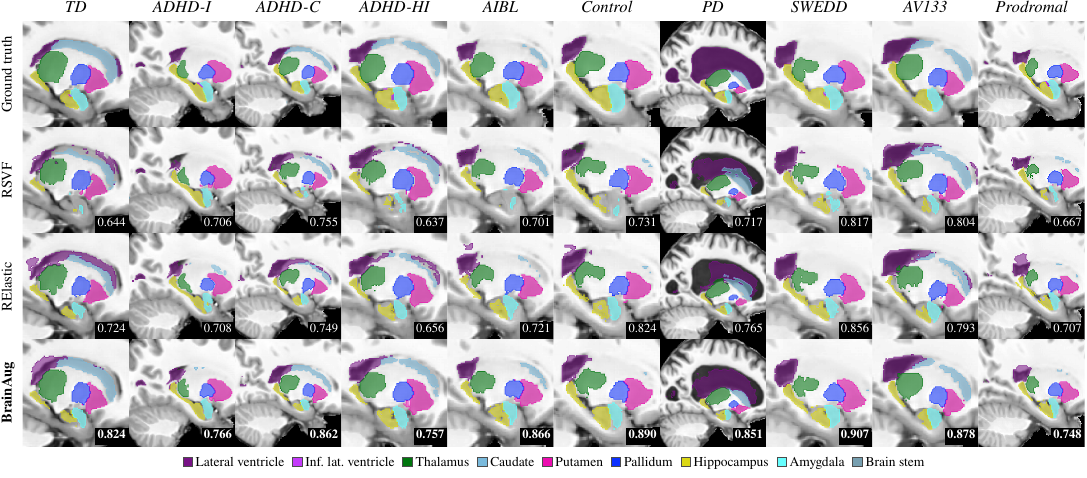}
    \caption{\textbf{Qualitative subcortical segmentation across clinical subgroups.}
    Sagittal views of $10$ subjects from ADHD200, AIBL, and PPMI, comparing ground truth with predictions from RSVF, RElastic, and \texttt{BrainAug}. Each column represents a clinical subgroup, with slice-level Dice over the displayed subcortical structures (best in bold). Colors distinguish the lateral and inferior lateral ventricles, thalamus, caudate, putamen, pallidum, hippocampus, amygdala, and brainstem. All models use the OASIS3-AD reference template.}
    \label{fig:supp_per_structure}
\end{figure*}

We extend the segmentation analysis of Sec.~\ref{sec:complement} with qualitative comparisons across $10$ clinical subgroups from ADHD200, AIBL, and PPMI. Figure~\ref{fig:supp_per_structure} compares RSVF, RElastic, and \texttt{BrainAug}, all trained on images synthesized from the same OASIS3-AD reference template. For each subgroup, we select one subject and visualize the subcortical structures in a sagittal view. Slices are chosen to capture multiple structures, favoring examples where \texttt{BrainAug} outperforms both baselines. These examples therefore illustrate characteristic differences rather than average performance.

Across all $10$ examples, \texttt{BrainAug} achieves the highest slice-level Dice ($0.748$--$0.907$), compared with RSVF ($0.637$--$0.817$) and RElastic ($0.656$--$0.856$). The improvements are particularly evident in the lateral ventricles and surrounding subcortical structures, where the anatomy-agnostic baselines exhibit boundary inaccuracies and incomplete segmentations. In contrast, \texttt{BrainAug} more closely follows the ground-truth anatomy across the different clinical subgroups. These qualitative observations complement the quantitative findings of Sec.~\ref{sec:complement}, illustrating the benefits of anatomically coordinated augmentation for subcortical segmentation across diverse anatomical presentations.

\section{Registration on Large Deformations}

Beyond segmentation, we investigate whether anatomically coordinated augmentation improves registration across substantial morphological differences. Learning-based registration can struggle when the anatomical variation encountered at inference exceeds that represented during training. SynthMorph~\cite{hoffmann2021synthmorph,hoffmann2024anatomy} addresses this limitation through synthetic training pairs generated using smooth random stationary velocity fields (SVFs). However, increasing the magnitude of these fields does not necessarily reproduce the coordinated regional expansion and contraction associated with large anatomical differences. In contrast, \texttt{BrainAug} explicitly controls regional log-volume changes through the sampling scale $\sigma$ and bound $c$, while graph-based correlations model dependencies between anatomical structures. This enables the synthesis of training pairs with pronounced, coordinated anatomical differences. We evaluate whether such training improves registration between children and older adults, where substantial differences in ventricular and subcortical anatomy pose a challenging test of deformation generalization.

\paragraph{Registration Training.}
Following SynthMorph~\cite{hoffmann2021synthmorph}, we train a 3D U-Net registration network using synthetic image pairs generated from the same $40$ reference segmentations. Let $(S_m^0,S_f^0)$ denote the sampled moving and fixed reference label maps. We generate synthetic segmentations $(S_m,S_f)$ by applying smooth SVF deformations followed by \texttt{BrainAug} transformations to the reference anatomies. We sample $70\%$ cross-subject pairs from distinct references, each independently deformed, and $30\%$ within-subject pairs sharing the same reference and initial SVF, followed by independently sampled or opposing \texttt{BrainAug} deformations. The resulting segmentations are converted into images $(I_m,I_f)$ through independently sampled contrast-agnostic intensity augmentation. For \texttt{BrainAug}, we sample $\sigma\sim\mathcal{U}(0.10,0.25)$ and $c\sim\mathcal{U}(0.5,1.0)$, using stronger variations ($\sigma\sim\mathcal{U}(0.20,0.25)$, $c\sim\mathcal{U}(0.8,1.0)$) for opposing deformations.

Given $(I_m,I_f)$, the registration network $f_\theta$ predicts a stationary velocity field $\mathbf{v}_\theta=f_\theta(I_m,I_f)$, which is integrated using seven scaling-and-squaring steps to obtain the diffeomorphic transformation $\phi_\theta=\exp(\mathbf{v}_\theta)$. We train the network to align the synthesized anatomical structures by minimizing a soft Dice loss between the warped moving and fixed segmentations, together with a velocity smoothness penalty:
\begin{equation}
\mathcal{L}_{\mathrm{reg}}(\theta)
=
\mathcal{L}_{\mathrm{Dice}}
\left(S_m\circ\phi_\theta,S_f\right)
+
\frac{\lambda}{|\Omega|}
\sum_{\mathbf{x}\in\Omega}
\left\|\nabla\mathbf{v}_\theta(\mathbf{x})\right\|_F^2,
\label{eq:reg_loss}
\end{equation}
where $\mathcal{L}_{\mathrm{Dice}}$ is computed on one-hot encoded segmentations, $\Omega$ denotes the image domain, and $\lambda=0.5$ controls velocity smoothness regularization. We train for $200{,}000$ iterations with a batch size of $1$, using Adam with an initial learning rate of $10^{-4}$.

\paragraph{Evaluation.} We evaluate all methods on $5{,}000$ bidirectional inter-subject registrations between $50$ ADHD200 subjects (7--26 years) and $50$ AIBL subjects (58--86 years), comprising all $2{,}500$ cross-dataset subject pairs in both directions. All methods receive the same affinely aligned moving and fixed images, and their predicted transformations are applied to the moving segmentations for evaluation. We assess $29$ anatomical structures: bilateral cerebral cortex, cerebral white matter, lateral and inferior lateral ventricles, cerebellar cortex and white matter, thalamus, caudate, putamen, pallidum, hippocampus, amygdala, and accumbens, together with the third and fourth ventricles and brainstem. Registration accuracy is quantified using Dice overlap, 95th-percentile Hausdorff distance (HD95), average symmetric surface distance (ASSD), and regional volume error $|\log(V_{\mathrm{moved}}/V_{\mathrm{fixed}})|$.  Structure-wise metrics are averaged across the $29$ regions, with mean and standard deviation reported over registrations.

\paragraph{Results.}  We compare against two official pretrained SynthMorph models. \textbf{SM-Def}, pretrained on FreeSurfer segmentations~\cite{fischl2002whole} with the same registration objective as ours, provides the primary comparison for assessing the contribution of anatomy-aware deformation sampling.  \textbf{SM-Joint}, pretrained on a composite dataset with a different objective, serves as an additional baseline under a broader training-data setting. 

\begin{figure*}[htbp]
    \centering
    \includegraphics[width=\linewidth]{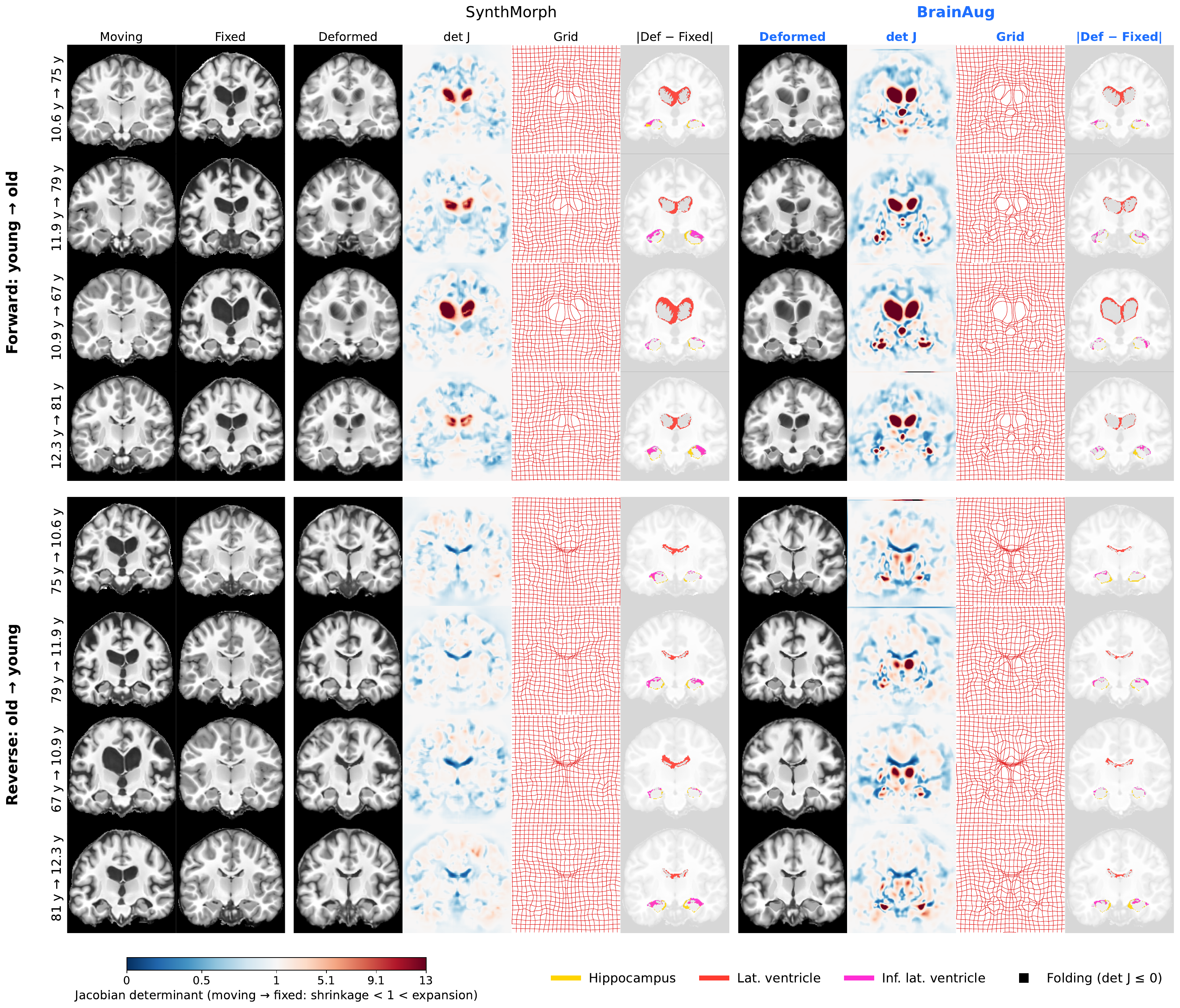}
    \caption{\textbf{Registration across large age-related anatomical differences.}
    Four ADHD200--AIBL subject pairs registered in both directions: young-to-old (top) and old-to-young (bottom), comparing SM-Def and \texttt{BrainAug}. For each method, we show the deformed moving image, Jacobian determinant ($\det J$), deformation grid, and segmentation disagreement with the fixed image for the hippocampus, lateral ventricle, and inferior lateral ventricle. The Jacobian maps visualize local expansion (red) and contraction (blue), with black indicating non-positive determinants. A shared color scale is used across methods and registration directions.}
    \label{fig:reg_hard}
\end{figure*}

Consistent with the quantitative results in the main paper, \texttt{BrainAug} achieves higher anatomical overlap, lower surface distances, and reduced regional volume mismatch compared with both pretrained SynthMorph models. Figure~\ref{fig:reg_hard} illustrates the differences between \texttt{BrainAug} and SM-Def across four child--older-adult subject pairs, registered in both directions. The Jacobian determinant maps reveal pronounced regional expansion around the ventricles during young-to-old registration and corresponding contraction in the reverse direction. Compared with SM-Def, \texttt{BrainAug} produces transformations that better accommodate these volume differences, with reduced segmentation disagreement around the lateral ventricles and neighboring subcortical structures.

\section{Runtime Analysis}

Tab.~\ref{tab:time} reports the per-volume runtime of each stage. Integration and warping are shared by all methods and cost the same (${\sim}28$\,ms and ${\sim}29$\,ms, respectively), so the methods differ only in how the field is generated. Despite encoding an anatomical prior and explicit volume targets, our velocity field is no more expensive to generate than an unstructured random field; it is in fact slightly cheaper ($22.9$ vs.\ ${\sim}26$\,ms). Two properties of the construction keep it cheap. First, the graph prior operates on ${\sim}40$ structure-level log-volume targets rather than on voxels, and is mapped to the volume by a single label lookup, enabling efficient voxel-wise mapping regardless of the number of anatomical structures. Second, our velocity is curl-free by construction: all three components are obtained as the gradient of a single scalar potential from one FFT-based Poisson solve, and its magnitude is bounded analytically by the volume targets. Random-field baselines, instead, sample and upsample each of the three components independently and rescale each one by its global maximum. Overall, the entire augmentation takes only $\mathbf{79.7}{\scriptstyle\,\pm2.1}$\,ms per volume on a single NVIDIA RTX PRO 6000 GPU, making it suitable for on-the-fly use during training.

\begin{table}[H]
  \centering
  \setlength{\tabcolsep}{3.5pt}
  \renewcommand{\arraystretch}{1.15}
  \resizebox{\columnwidth}{!}{%
  \begin{tabular}{@{}lcccc@{}}
    \toprule
    Method & Field\,$\downarrow$ & Integration\,$\downarrow$ & Warp\,$\downarrow$ & Total\,$\downarrow$ \\
    \midrule
    RElastic
      & \underline{$25.9$}${\scriptstyle\,\pm1.1}$
      & \underline{$28.1$}${\scriptstyle\,\pm0.6}$
      & $29.1{\scriptstyle\,\pm1.2}$
      & \underline{$83.2$}${\scriptstyle\,\pm2.5}$ \\
    RSVF
      & $26.2{\scriptstyle\,\pm1.1}$
      & \underline{$28.1$}${\scriptstyle\,\pm0.6}$
      & $29.1{\scriptstyle\,\pm1.2}$
      & $83.4{\scriptstyle\,\pm2.6}$ \\
    \midrule
    \rowcolor{oursrow}
    \textbf{Ours}
      & $\mathbf{22.9}{\scriptstyle\,\pm0.7}$
      & $\mathbf{27.8}{\scriptstyle\,\pm0.6}$
      & $29.1{\scriptstyle\,\pm1.2}$
      & $\mathbf{79.7}{\scriptstyle\,\pm2.1}$ \\
    \bottomrule
  \end{tabular}}
  \caption{Per-volume runtime (ms) on a single NVIDIA RTX PRO 6000 GPU. Mean\,$\pm$\,std over $1000$ runs on a $182{\times}218{\times}182$ volume. All methods integrate their field (scaling and squaring) and share the same warp. \textbf{Bold}: best; \underline{underline}: second best.}
  \label{tab:time}
\end{table}

\end{document}